\documentclass[lettersize,journal]{IEEEtran}
\usepackage{amsmath,amsfonts}
\usepackage{array}
\usepackage[caption=false,font=normalsize,labelfont=sf,textfont=sf]{subfig}
\usepackage{textcomp}
\usepackage{stfloats}
\usepackage{url}
\usepackage{verbatim}
\usepackage{graphicx}
\usepackage{cite}
\usepackage[numbers]{natbib}   
\usepackage{tabularx}
\usepackage{adjustbox}
\usepackage{CJKutf8}
\usepackage{amssymb}
\usepackage{amsmath}
\usepackage{multirow}
\usepackage{array}
\usepackage[flushleft]{threeparttable}
\usepackage{xcolor}
\usepackage{longtable}
\usepackage{booktabs}
\usepackage{bm}
\usepackage{float}
\usepackage{makecell}
\usepackage[table]{xcolor}
\usepackage{booktabs}
\usepackage{multirow}       
\usepackage{array}
\usepackage{amsmath, amssymb}
\usepackage{graphicx}
\usepackage{subcaption}
\usepackage{tikz}
\usetikzlibrary{tikzmark}

\usepackage{algorithm}
\usepackage{algpseudocode} 

\begin{document}
\title{RLGPA: A Reinforcement Learning-Guided Granularity Planning Attack for Decision-Based Sparse Adversarial Examples}
\author{Chao Zhou, Shenghua He, Jiaxin Mo, Qiangguo Yu, Fengchuang Xing
\thanks{Chao Zhou, Jiaxin Mo, Qiangguo Yu are with the School of Electronic Information, Huzhou College, Huzhou, 313000, China (e-mail: zhouchaodzkd@e.gzhu.edu.cn; zjhzumojiaxin@zjhzu.edu.cn; yuqg@zjhzu.edu.cn).}
\thanks{Shenghua He is with the School of Electronic Science and Engineering, South China Normal University, Foshan, 528225, China (e-mail: heshenghua@m.scnu.edu.cn).}
\thanks{Fengchuang Xing is with the School of Physics and Information Engineering, Guangdong University of Education, Guangzhou, 510303, China (e-mail: xfchuang@e.gzhu.edu.cn).}
\thanks{Shenghua He and Chao Zhou are equally contributed to this work.}
\thanks{Corresponding author: Shenghua He.}}
\maketitle

\begin{abstract}
Decision-based sparse attacks must reduce perturbation support using only the victim's predicted label. The best restoration strategy varies as support contracts: coarse removal works early, while tail states need finer, history-aware tests. We propose RLGPA, an online reinforcement learning method that schedules support-reduction executors and their granularity under a hard query budget. A masked Dueling Double DQN selects among seven stable policy--scale actions; each action produces support-relative restoration groups that are rebuilt after every macro decision. A feasibility-preserving executor commits only adversarial restorations and feeds support gain and query cost back to the planner. Under the matched top-1-label oracle, RLGPA reaches 61\% targeted ASR at sparsity 0.01 on CIFAR-10 (vs. 25\% SparseEvo, 0\% Pointwise), and 75.5\%/73.0\% on ResNet50/ViT at sparsity 0.005 (vs. 29.0\%/15.5\%). A fixed-probe test reduces median \(L_0\) from 165 to 23 after 40 images without any update. State-blind and state-shuffled controls degrade significantly, confirming that the learned policy uses attack state; the main gains come from state-conditioned executor scheduling and dynamic repartition. The source code is publicly available at https://github.com/Acui001/RLGPA.git.
\end{abstract}


\section{Introduction}
Deep neural networks achieve strong performance across a wide range of perception tasks, yet their vulnerability to adversarial examples, which are inputs crafted by adding small, often human-imperceptible perturbations that cause incorrect predictions, introduces significant security risks to deep learning-based visual detection tasks \cite{11145157, 10741570, szegedy2013intriguing, ZHOU2024103710}. 
This vulnerability has motivated a large body of work on adversarial  attacks \cite{11595869,11298158,ZHOU2025117241,zhou2025query}, which differ in the threat model they assume and in the constraints they place on the perturbation. Among these, sparse adversarial attacks are particularly relevant to security-critical settings, because they modify only a limited number of input dimensions and therefore correspond more closely to the practical constraint that an attacker can manipulate only a small part of the input. 

Sparse adversarial attacks seek an adversarial input whose perturbation involves only a small subset of image pixels. The underlying minimum-$L_0$ adversarial problem is NP-hard \cite{vo2022spEVo}. In a decision-based black-box setting, the attacker observes only the top-1 predicted label and cannot access gradients, logits, or confidence scores. As a result, exact support minimization, i.e., minimizing the set of modified pixels, becomes a query-bounded combinatorial search. Starting from a feasible adversarial point, Pointwise attack \cite{schott2019pointwise} and SparseEvo attack \cite{vo2022spEVo} approach this problem through perturbation erasure: they ask which active coordinates can be restored to the original image without losing adversarial feasibility. Pointwise greedily tests restorations, whereas SparseEvo evolves binary support masks. Their results show that support erasure is central to decision-based sparse attacks, but they also expose a scheduling problem: the best restoration ordering and group size need not remain fixed throughout an attack.

This scheduling problem arises because the search has a pronounced stage dependence. Large groups can remove support quickly when many perturbed coordinates are redundant. However, as the support shrinks, the same removals cross the decision boundary more often, and the attack benefits from finer groups or from empirical knowledge about previous restoration outcomes. A fixed schedule cannot condition on current progress, stalls, acceptance rate, or remaining budget. Conversely, directly assigning reinforcement-learning actions to temporary pixel groups is ill posed: dynamic repartition changes the coordinates represented by an action index, making replay transitions semantically unstable.

To address this problem, we propose RLGPA, a reinforcement learning-guided granularity planning attack, where reinforcement learning (RL) schedules the executor policy and granularity. Instead of selecting a pixel-group index, the agent selects a stable macro-action that composes an executor policy with a relative scale. The executor instantiates that action on the current support, tests a bounded number of restorations, and commits only adversarial candidates. The next macro decision is therefore made after observing an actual support reduction and query cost. Because Primary, Safe, and Memory retain their meanings after repartition and across images, transitions can be reused by an online Dueling Double DQN.

RLGPA uses support-relative granularity. If $L_t$ coordinates remain perturbed, an action selects a group size proportional to $L_t$ rather than to the original image dimension. This produces the same seven action semantics on CIFAR-10 and ImageNet despite their very different resolutions. Dynamic repartition then rebuilds groups around the newly committed adversarial point. The DQN observes the optimization phase, current granularity, recent gains and acceptances, stall counts, remaining budget, recent action identity, action-specific effectiveness, and the valid-action mask.

Our contributions are as follows:
\begin{itemize}
    \item We formulate decision-based sparse attack refinement as online reinforcement learning over stable policy--granularity macro-actions. This abstraction avoids attaching persistent Q-values to temporary support groups and makes cross-decision and cross-image replay meaningful.
    \item We develop a support-relative, dynamically repartitioned executor that integrates three complementary restoration policies: perturbation-ascending, magnitude-aware, and outcome-memory. Through hard-label verification and a commit-or-rollback execution scheme, this executor monotonically reduces the support while strictly preserving adversarial feasibility.
    \item We provide matched causal evidence for the learning component. Frozen fixed-probe evaluation, state-blind and state-shuffled controls, random and fixed schedulers, lifecycle controls, and component ablations distinguish learned state-conditioned scheduling from executor strength and favorable initialization.
    \item We evaluate targeted and untargeted attacks on CIFAR-10 and ImageNet with ResNet50 and Vision Transformer (ViT) victims. Under an oracle-matched top-1-label protocol, RLGPA achieves improved Query--Sparsity and ASR--Sparsity performance relative to comparable decision-based sparse attacks.
\end{itemize}

\section{Related Work}

\subsection{Decision-Based Black-Box Attacks}
Hard-label attacks observe only the top-1 decision. Boundary Attack introduced random walks along the decision boundary \cite{brendel2018decision}; HopSkipJump improved this process with geometric steps and binary search \cite{chen2019hopskipjump}, while Yan \textit{et al.} learned query placement with reinforcement learning \cite{yan2021policy}. Other methods construct continuous boundary-distance objectives: Ilyas \textit{et al.} estimated a query-limited surrogate \cite{ilyas2018black}, Opt-attack minimized distance to the boundary \cite{cheng2019query}, and Sign-OPT and Sign-OPT+ improved zeroth-order sign optimization \cite{ChengSCC0H20,ran2022sign}. AdvViT further exploits transformer structure for efficient black-box attacks \cite{zhou2025query}. 


\subsection{Sparse Adversarial Attacks}
Sparse attacks restrict nonzero perturbations to a small $L_0$ support, directly modeling settings in which only a limited number of input coordinates can be changed. 

Under hard-label feedback, Pointwise greedily restores coordinates while preserving adversarial feasibility \cite{schott2019pointwise}, whereas SparseEvo evolves support masks and performs strongly on CNN and transformer victims \cite{vo2022spEVo}. Both reduce an existing adversarial support using top-1 queries; their main difference is the restoration controller.

Richer score-based oracles support Sparse-RS random search \cite{croce2022sparse}, Bayesian sparsity search in BruSLeAttack \cite{vo2024brusleattack}, and joint sparsity--magnitude optimization in SA-MOO \cite{williams2023black}. Related score-based attacks include bandit/prior-guided gradient estimation \cite{ilyas2018prior,ilyas2019black}, SimBA \cite{guo2019simple}, Square Attack \cite{andriushchenko2020square}, and the one-pixel attack \cite{su2019one}. Because a score vector contains more information than a top-1 label, these methods are relevant adjacent work but not oracle-matched query baselines. 

White-box PGD$_0$ \cite{croce2019sparse} uses iterative magnitude pruning, while $\sigma$-zero \cite{cina2025sigmazero} smooths sparse projection. They provide ASR--Sparsity references but not Query--Sparsity baselines because they use gradients rather than victim queries. Their broader lineage includes L-BFGS \cite{szegedy2013intriguing}, FGSM \cite{GoodfellowSS14}, JSMA \cite{papernot2016limitations}, DeepFool \cite{moosavi2016deepfool}, C\&W \cite{carlini2017towards}, and PGD \cite{madry2017towards}; transfer attacks instead craft examples on a surrogate \cite{liu2016delving}. 

\subsection{Reinforcement Learning for Query Planning}
Unlike gradient-based FGSM, PGD, and C\&W \cite{GoodfellowSS14,madry2017towards,carlini2017towards}, RL can schedule black-box queries from sequential feedback. Prior work learned query placement \cite{yan2021policy} or optimized adversarial-patch location and perturbation \cite{wei2022simultaneously}; other black-box optimizers include NES \cite{ilyas2018black} and AutoZOOM \cite{tu2019autozoom}. RLGPA instead uses standard DQN, Double-DQN, and Dueling components \cite{mnih2015human,van2016deep,wang2016dueling} to schedule stable support-reduction operators online. Its distinction is therefore the state-conditioned action abstraction, not a new RL backbone or a claim of broad policy generalization \cite{hafner2025mastering,NEURIPS2023_a7a7c0c9}.

\section{Method}
RLGPA formulates decision-based sparse attack as online, query-bounded support reduction. A compact state-conditioned planner selects a restoration policy and relative granularity, while PSSE executes and verifies every restoration; the following subsections define the objective and the resulting closed loop.

\subsection{Problem Formulation}

Let \(f:[0,1]^{C\times H\times W}\!\to\!\{1,\ldots,K\}\) return only the top-1 class. For an image \(\mathbf{x}\) with \(y=f(\mathbf{x})\), the untargeted and targeted feasibility predicates \(\mathcal{A}_{\mathrm{u}}\) and \(\mathcal{A}_{\mathrm{t}}\) for a candidate image \(\mathbf{z}\) are:
\begin{equation}
\mathcal{A}_{\mathrm{u}}(\mathbf{z})=\mathbb{I}[f(\mathbf{z})\neq y],
\qquad
\mathcal{A}_{\mathrm{t}}(\mathbf{z})=\mathbb{I}[f(\mathbf{z})=y^{\star}],
\label{eq:feasibility}
\end{equation}
where \(\mathbb{I}[\cdot]\) is the indicator function and \(y^{\star}\neq y\) is the prescribed target label.

RLGPA measures support over scalar tensor coordinates:
\begin{equation}
\begin{aligned}
\mathcal{I}_{\mathrm{ch}}&=\{(c,h,w):c\in[C],h\in[H],w\in[W]\},\\
\Omega(\mathbf{z},\mathbf{x})&=\{i\in\mathcal{I}_{\mathrm{ch}}:z_i\neq x_i\},\\
L(\mathbf{z},\mathbf{x})&=|\Omega(\mathbf{z},\mathbf{x})|,\\
s(\mathbf{z},\mathbf{x})&=L(\mathbf{z},\mathbf{x})/(CHW),
\end{aligned}
\label{eq:channel-support}
\end{equation}
where \([n]=\{1,\ldots,n\}\), \(i=(c,h,w)\) indexes one scalar entry, \(x_i\) and \(z_i\) are the corresponding values, \(\Omega\) is the active support, \(L\) its cardinality, and \(s\) normalized sparsity. If all \(C\) channels share a spatial mask of size \(L_{\mathrm{pix}}\), then \(L=C L_{\mathrm{pix}}\) and \(s=L_{\mathrm{pix}}/(HW)\), so channel-coordinate and spatial-pixel representations yield the same normalized sparsity.

Starting from a feasible adversarial image \(\mathbf{z}_0\), RLGPA restores coordinates rather than estimating new amplitudes. Let \(\Omega_0=\Omega(\mathbf{z}_0,\mathbf{x})\) and \(\mathbf{r}\in\{0,1\}^{|\Omega_0|}\) be its initial support and a binary retention vector over that support. The induced candidate is
\begin{equation}
z_i(\mathbf{r})=
\begin{cases}
x_i+r_i(z_{0,i}-x_i),& i\in\Omega_0,\\
x_i,&i\notin\Omega_0,
\end{cases}
\label{eq:retention-candidate}
\end{equation}
where \(z_i(\mathbf{r})\) is coordinate \(i\) of \(\mathbf{z}(\mathbf{r})\), \(z_{0,i}\) and \(x_i\) are its initial adversarial and clean values, and \(r_i=1\) retains whereas \(r_i=0\) restores the coordinate. Coordinates outside \(\Omega_0\) already equal \(\mathbf{x}\). The query-bounded restoration problem is
\begin{equation}
\min_{\mathbf{r}\in\{0,1\}^{|\Omega_0|}}\;\|\mathbf{r}\|_0
\quad\mathrm{s.t.}\quad
\mathcal{A}(\mathbf{z}(\mathbf{r}))=1,
\qquad Q\leq B,
\label{eq:erasure-objective}
\end{equation}
where \(\|\mathbf{r}\|_0\) counts retained coordinates, \(Q\) is the cumulative number of top-1 victim queries, and \(B\) is the query budget. Each candidate needs one hard-label query, including the initial feasibility check \(Q_0=1\). The \(2^{|\Omega_0|}\) restoration search is a restricted NP-hard minimum-\(L_0\) problem \cite{vo2022spEVo}; RLGPA is a query-efficient heuristic, not a global optimizer.

\subsection{RLGPA Framework}

\label{sec:overview}
 
 \begin{figure*}[tp]
\centering
\includegraphics[width=\linewidth]{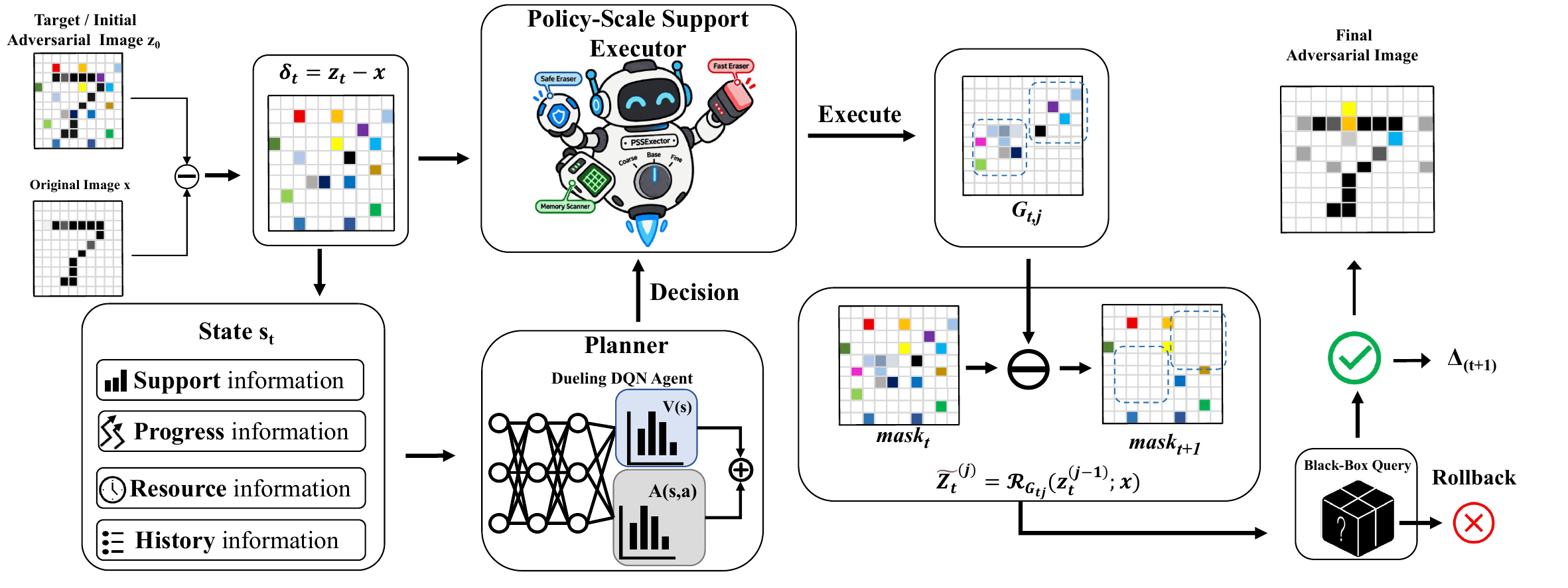}
\caption{{Overview of RLGPA. The planner selects a policy--scale action from the current attack state, PSSE performs hard-label-verified support restoration, and the observed support reduction and query cost are returned for online updating.}}
\label{fig:rlgpa-main}
\end{figure*}

Figure~\ref{fig:rlgpa-main} summarizes the closed loop. A compact state records support progress, remaining budget, and recent operator outcomes; a masked Dueling Double DQN selects a policy--scale macro-action; PSSE rebuilds and tests restoration groups; and the observed support reduction and query cost update the state and planner. The policy controls group construction and ordering, while the scale controls relative granularity. The planner therefore schedules stable operator semantics rather than pixel identities, and every accepted restoration is independently verified by the hard-label oracle.

\subsection{Attack-State Representation}
\label{sec:state-repr}

The state describes search progress rather than raw pixels. Let \(L_0=L(\mathbf{z}_0,\mathbf{x})\) and \(L_t=L(\mathbf{z}_t,\mathbf{x})\) be the initial and current support sizes, \(Q_t\) the cumulative queries before decision \(t\), \(H_t\) the number of consecutive preceding zero-gain decisions, and \(\widehat N_t\) the group-count descriptor carried from the preceding decision. At \(t=0\), \(\widehat N_t\) is initialized from the clipped initial partition; subsequently it is updated by~\eqref{eq:feedback-state-update}. The normalized progress variables are
\begin{equation}
\ell_t=\frac{L_t}{\max(1,L_0)},
\;
\nu_t=\frac{\widehat N_t}{\max(1,L_t)},
\;
\varrho_t=\frac{B-Q_t}{B},
\label{eq:state-progress}
\end{equation}
where \(\ell_t\), \(\nu_t\), and \(\varrho_t\) are the remaining-support, carried-group-density, and remaining-budget ratios, and \(\max(1,\cdot)\) prevents division by zero. The continuous feature vector is
\begin{equation}
\begin{split}
\boldsymbol{\xi}_t = \big[ &\ell_t,\ \nu_t,\ \min(1, H_t/5),\ \operatorname{clip}(\zeta_{t-1},0,1), \\
&\operatorname{clip}(\alpha_{t-1},0,1),\ \operatorname{clip}(\varrho_t,0,1) \big],
\end{split}
\label{eq:numeric-state}
\end{equation}
where \(\boldsymbol\xi_t\in\mathbb{R}^{6}\), \(\zeta_{t-1}=\Delta L_{t-1}/\max(1,L_{t-1})\) is the preceding relative support reduction, \(\Delta L_{t-1}=L_{t-1}-L_t\), \(\alpha_{t-1}\) is the preceding acceptance ratio, and \(\operatorname{clip}(v,0,1)=\min(1,\max(0,v))\); initially \(\zeta_{-1}=\alpha_{-1}=0\). The categorical vector \(\mathbf{c}_t\in\{0,1\}^{21}\) concatenates one-hot bins for: phase \(\phi_t\), which is early for \(\ell_t>0.35\), middle for \(0.08<\ell_t\leq0.35\), and late otherwise; granularity over \([0,0.25)\), \([0.25,0.75)\), \([0.75,0.95)\), and \([0.95,1]\) of \(\nu_t\); three bins each for stall \(H_t\in\{0,1,\geq2\}\), gain \(\zeta_{t-1}\) at \(0\) and \(0.05\), acceptance \(\alpha_{t-1}\) at \(0\) and \(0.25\), and budget \(\varrho_t\) at \(0.15\) and \(0.5\); plus a binary legacy-safety flag fixed off in the official configuration.

Let \(\mathcal{W}_t\) contain the latest \(W=8\) within-image outcomes and \(\mathcal{W}_t(a)=\{\tau\in\mathcal{W}_t:a_\tau=a\}\) the subset generated by action \(a\). Its recent gain efficiency and hit rate are
\begin{equation}
\begin{aligned}
& e_t(a)=\min\!\left(1,
\frac{\sum_{\tau\in\mathcal{W}_t(a)}\Delta L_\tau}
{\max\!\left(1,\sum_{\tau\in\mathcal{W}_t(a)}u_\tau\right)}\right), \\
& h_t(a)=\frac{\sum_{\tau\in\mathcal{W}_t(a)}\mathbb{I}[\Delta L_\tau>0]}
{\max(1,|\mathcal{W}_t(a)|)},
\end{aligned}
\label{eq:action-history}
\end{equation}
where \(\tau\) indexes a stored decision, \(\Delta L_\tau\) and \(u_\tau\) are its support reduction and query use, \(e_t(a)\) is clipped accumulated gain per query, and \(h_t(a)\) its positive-outcome fraction; both are zero for unseen actions. Mean per-decision efficiency is
\begin{equation}
\bar e_t(a)=\frac{1}{\max(1,|\mathcal{W}_t(a)|)}
\sum_{\tau\in\mathcal{W}_t(a)}
\frac{\Delta L_\tau}{\max(1,u_\tau)},
\label{eq:recent-best}
\end{equation}
where \(\bar e_t(a)=0\) when \(\mathcal{W}_t(a)\) is empty. The vector \(\mathbf{o}_t\in\{0,1\}^{M+1}\) marks \(\arg\max_a\bar e_t(a)\) among actions with positive accumulated gain, with an extra ``none'' slot otherwise. Let \(\boldsymbol\mu_t\in\{0,1\}^{M}\) be the valid-action mask, with \(\mu_t(a)=1\) iff action \(a\) may be selected. The complete state is
\begin{align}
\mathbf{s}_t&=\operatorname{concat}\!\left(
\mathbf{c}_t,\mathbf{o}_t,\boldsymbol\xi_t,
\{e_t(a),h_t(a)\}_{a=1}^{M},\boldsymbol\mu_t\right),\nonumber\\
d_s&=21+(M+1)+6+2M+M=28+4M,
\label{eq:state-dim}
\end{align}
where \(M\) is the action count, \(\{e_t(a),h_t(a)\}_{a=1}^{M}\) concatenates both statistics for every action, and \(d_s\) is the state dimension. With \(M=7\), \(d_s=56\). Actions with \(\kappa>1\) are masked when \(\nu_t\geq0.95\), i.e., at one-coordinate granularity. Histories and coordinate memories reset per image, while declared planner weights and replay may persist. Thus the state schedules support-reduction operators but contains neither support geometry nor semantic pixel identities.

\subsection{Online RL-Based Macro-Action Planner}
\label{sec:dqn}

RLGPA models operator scheduling as the finite-horizon discounted process \(\mathcal{P}=(\mathcal{S},\mathcal{M},P,r,\gamma)\), where \(\mathcal{S}\), \(\mathcal{M}\), \(P(\mathbf{s}_{t+1}\mid\mathbf{s}_t,a_t)\), \(r\), and \(\gamma\in[0,1)\) are the scheduler-state space, macro-action set, executor-induced transition, reward, and discount factor. Because \(\mathbf{s}_t\) omits image geometry, this is an observable scheduling approximation rather than an exact Markov claim. Each action is
\begin{equation}
a_t=(p_t,\kappa_t)\in\mathcal{M},
\quad
p_t\in\{\mathrm{Primary},\mathrm{Safe},\mathrm{Memory}\},
\label{eq:macro-action}
\end{equation}
where \(a_t\) is the decision at step \(t\), \(p_t\) selects an executor, and \(\kappa_t>0\) its relative scale. The seven coordinate-independent policy--scale pairs in Table~\ref{tab:actions} retain their meanings after repartition and across images.

The masked Dueling Double DQN decomposes action value as
\begin{equation}
Q_\theta(\mathbf{s},a)=V_\theta(\mathbf{s})+A_\theta(\mathbf{s},a)
-\frac{1}{M}\sum_{a'=1}^{M}A_\theta(\mathbf{s},a'),
\label{eq:dueling-q}
\end{equation}
where \(Q_\theta(\mathbf{s},a)\) is the estimated return, \(\theta\) the online parameters, \(V_\theta(\mathbf{s})\) and \(A_\theta(\mathbf{s},a)\) the value and advantage streams, and \(a'\) indexes all \(M\) actions. An \(\epsilon\)-greedy policy samples only \(\mathcal{M}_t=\{a\in\mathcal{M}:\mu_t(a)=1\}\), with invalid values masked to \(-\infty\). For terminal flag \(d_t\), the Double-DQN update is
\begin{align}
a_t^\star&=\arg\max_{a\in\mathcal{M}_{t+1}}Q_\theta(\mathbf{s}_{t+1},a),\\
y_t&=r_t+\gamma(1-d_t)Q_{\bar\theta}(\mathbf{s}_{t+1},a_t^\star),\\
\mathcal{L}(\theta)&=\mathbb{E}_{\mathcal{D}}\!\left[
\operatorname{Huber}\!\left(Q_\theta(\mathbf{s}_t,a_t)-y_t\right)\right],
\label{eq:ddqn-target}
\end{align}
where \(\mathcal{M}_{t+1}\) is the next valid set, \(a_t^\star\) its online-network maximizer, \(r_t\) the observed reward, \(y_t\) the target, \(\bar\theta\) the target-network parameters, \(\mathcal{D}\) replay memory, and \(\operatorname{Huber}\) the smooth-\(L_1\) loss. The stored next mask defines \(\mathcal{M}_{t+1}\); \(d_t\in\{0,1\}\) removes bootstrapping at terminal transitions and image boundaries.

Both networks use a 128--128 ReLU trunk and separate 64-unit heads. Adam uses learning rate \(10^{-3}\), \(\gamma=0.8\), batch size 64, replay capacity 20,000, warm-up 128, gradient clipping 5, and target synchronization every 250 updates. Exploration decays \(0.15\to0.02\) by factor 0.995 per decision and resets per image. Continual-online evaluation preserves weights and replay; frozen evaluation disables exploration and all learning-state mutation, separating executor capability from policy learning.

\begin{table}[t]
\centering
\caption{Canonical RLGPA macro-actions. Policy--scale semantics remain fixed when the current support is repartitioned.}
\label{tab:actions}
 \setlength{\tabcolsep}{3pt}
\begin{tabular}{llll}
\toprule
Action & Executor & Ordering & $\kappa$ \\
\midrule
Coarse Primary & Primary & perturbation ascending & 0.5 \\
Base Primary & Primary & perturbation ascending & 1.0 \\
Coarse Safe & Safe & mean magnitude descending & 0.5 \\
Base Safe & Safe & mean magnitude descending & 1.0 \\
Fine Safe & Safe & mean magnitude descending & 2.0 \\
Base Memory & Memory & restoration memory descending & 1.0 \\
Fine Memory & Memory & restoration memory descending & 2.0 \\
\bottomrule
\end{tabular}
\end{table}

\subsection{Policy-Scale Support Executor}
\label{sec:psse}

PSSE instantiates \(a_t=(p_t,\kappa_t)\) on the verified incumbent \(\mathbf{z}_t\), whose active support and size are \(\Omega_t=\Omega(\mathbf{z}_t,\mathbf{x})\) and \(L_t=L(\mathbf{z}_t,\mathbf{x})=|\Omega_t|\). Given base erase ratio \(\rho>0\), it computes
\begin{equation}
\begin{aligned}
\eta_t&=\rho/\kappa_t,\\
g_t&=\min\!\left(L_t,\max\!\left(g_{\min},
\left\lceil L_t\eta_t\right\rceil\right)\right),\\
J_t&=\left\lceil L_t/g_t\right\rceil,
\end{aligned}
\label{eq:relative-groups}
\end{equation}
where \(\eta_t\), \(g_t\), and \(J_t\) are the requested support fraction per group, integer target group size, and group count; \(g_{\min}\) is the size lower bound and \(\lceil\cdot\rceil\) the ceiling operator. We use \(\rho=0.002\), \(g_{\min}=1\), and \(\kappa_t\in\{0.5,1,2\}\) for coarse, base, and fine scales. The realized ratio \(\widehat\eta_t=g_t/L_t\) may differ from \(\eta_t=\rho/\kappa_t\) because of integer clipping.

Let \(g_t^{\mathrm{lo}}=\lfloor L_t/J_t\rfloor\) be the smaller balanced size and \(R_t^{\mathrm{bal}}=L_t-J_tg_t^{\mathrm{lo}}\) the number of groups receiving one extra coordinate. The partition \(\{G_{t,j}\}_{j=1}^{J_t}\) satisfies
\begin{equation}
\begin{aligned}
|G_{t,j}| &= \begin{cases}
g_t^{\mathrm{lo}}+1, & 1\leq j\leq R_t^{\mathrm{bal}},\\
g_t^{\mathrm{lo}}, & R_t^{\mathrm{bal}}<j\leq J_t,
\end{cases} \\
& G_{t,j}\cap G_{t,k}=\varnothing \ (j\neq k), \quad
\bigcup_{j=1}^{J_t}G_{t,j}=\Omega_t,
\end{aligned}
\label{eq:balanced-partition}
\end{equation}
where \(\lfloor\cdot\rfloor\) is the floor operator, \(G_{t,j}\subseteq\Omega_t\) is group \(j\), and \(j,k\in\{1,\ldots,J_t\}\). The groups are exhaustive, disjoint, and directly generated by PSSE. Rebuilding them from committed support before every macro-action preserves policy--scale semantics without assigning Q-values to temporary group indices.

Policy \(p_t\) controls group construction and order. For \(i\in\Omega_t\), let \(\omega_i=|z_{t,i}-x_i|\) be perturbation magnitude. \textbf{Primary} shuffles ties, sorts coordinates by increasing \(\omega_i\), and chunks the sequence. \textbf{Safe} randomly groups coordinates and ranks groups by decreasing
\begin{equation}
S_{\mathrm{mag}}(G)=\frac{1}{|G|}\sum_{i\in G}\omega_i,
\label{eq:safe-score}
\end{equation}
where \(G\subseteq\Omega_t\), \(|G|\) is its size, and \(S_{\mathrm{mag}}(G)\) its mean magnitude. \textbf{Memory} maintains image-local accepted and rejected group attributions \((n_i^+,n_i^-)\) and ranks groups by
\begin{equation}
m_i=\frac{n_i^++1}{n_i^++n_i^-+2},
\qquad
S_{\mathrm{mem}}(G)=\frac{1}{|G|}\sum_{i\in G}m_i,
\label{eq:memory}
\end{equation}
where \(n_i^+\) and \(n_i^-\) count accepted and rejected tests containing \(i\), \(m_i\in(0,1)\) is the Beta\((1,1)\)-smoothed attribution, and \(S_{\mathrm{mem}}(G)\) its group mean. Testing \(G\) increments \(n_i^+\) for every \(i\in G\) on acceptance and \(n_i^-\) otherwise. These per-image executor statistics are ordering scores, not identified coordinate-level causal probabilities or cross-image DQN state.

Set \(\mathbf{z}_t^{(0)}=\mathbf{z}_t\). For group \(G_{t,j}\), the restoration operator and candidate are
\begin{equation}
[\mathcal{R}_{G}(\mathbf{z};\mathbf{x})]_i=
\begin{cases}
x_i,&i\in G,\\
z_i,&i\notin G.
\end{cases}
\;
\widetilde{\mathbf{z}}_t^{(j)}=
\mathcal{R}_{G_{t,j}}(\mathbf{z}_t^{(j-1)};\mathbf{x}),
\label{eq:restore-operator}
\end{equation}
where \(\mathcal{R}_G(\mathbf{z};\mathbf{x})\) copies \(x_i\) into scalar coordinates \(i\in G\) and retains \(z_i\) elsewhere, \(\mathbf{z}_t^{(j-1)}\) is the latest verified incumbent, and \(\widetilde{\mathbf{z}}_t^{(j)}\) is the unverified candidate. One victim query returns \(\chi_{t,j}=\mathcal{A}(\widetilde{\mathbf{z}}_t^{(j)})\in\{0,1\}\), after which PSSE applies
\begin{equation}
\begin{aligned}
\mathbf{z}_t^{(j)} &= 
\begin{cases}
\widetilde{\mathbf{z}}_t^{(j)}, & \chi_{t,j}=1,\\
\mathbf{z}_t^{(j-1)}, & \chi_{t,j}=0,
\end{cases} \\
L(\mathbf{z}_t^{(j)},\mathbf{x}) &= 
L(\mathbf{z}_t^{(j-1)},\mathbf{x})-\chi_{t,j}|G_{t,j}|,
\end{aligned}
\label{eq:commit-rollback}
\end{equation}
where \(\mathbf{z}_t^{(j)}\) is the verified incumbent after test \(j\); \(\chi_{t,j}=1\) commits the candidate and \(\chi_{t,j}=0\) rolls it back. Since tested groups are active and disjoint, every incumbent remains feasible and support never increases. A macro-action executes
\begin{equation}
u_t=\min(q_c,J_t,B-Q_t),
\qquad q_c=32,
\label{eq:query-chunk}
\end{equation}
tests, where \(u_t\) is both the number of tested groups and consumed victim queries, \(q_c=32\) is the per-decision cap, \(J_t\) the available group count, and \(B-Q_t\) the remaining budget. It returns
\begin{equation}
\begin{aligned}
\mathbf{z}_{t+1} &= \mathbf{z}_t^{(u_t)}, \\
\Delta L_t &= L_t - L_{t+1}
= \sum_{j=1}^{u_t} \chi_{t,j} |G_{t,j}|, \\
\alpha_t &= \frac{1}{\max(1,u_t)}
\sum_{j=1}^{u_t} \chi_{t,j},
\end{aligned}
\label{eq:macro-outcome}
\end{equation}
where \(\mathbf{z}_{t+1}=\mathbf{z}_t^{(u_t)}\) is the returned incumbent, \(L_{t+1}=L(\mathbf{z}_{t+1},\mathbf{x})\), \(\Delta L_t\) the accepted support reduction, and \(\alpha_t\) the accepted-test ratio. Thus PSSE performs feasibility-preserving support reduction; RL only schedules its executor and scale.

\subsection{Query-Aware Feedback}
\label{sec:feedback}
\label{sec:reward-def}

PSSE returns \((\Delta L_t,u_t,\alpha_t)\), acceptance bits \(\{\chi_{t,j}\}_{j=1}^{u_t}\), and updated coordinate memory. Let \(\bar g_t=L_t/J_t\) be mean planned group size, \(H_t\) the pre-action stall length, and \(\phi_t\) the phase determined by \(\ell_t=L_t/L_0\). The reward is
\begin{align}
r_t={}&w_{\phi_t}\frac{\Delta L_t}{\max(1,u_t\bar g_t)}
+\beta_{\mathrm{tail}}\mathbb{I}[\Delta L_t>0\wedge\phi_t=\mathrm{late}]\nonumber\\
&+\beta_{\mathrm{stall}}\mathbb{I}[\Delta L_t>0\wedge H_t\geq2]\nonumber\\
&+c_0\mathbb{I}[\Delta L_t=0] -\lambda_q\frac{u_t}{B},
\label{eq:reward}
\end{align}
where \(r_t\) is scalar feedback, \(w_{\phi_t}\) the phase weight, \(\beta_{\mathrm{tail}}\) and \(\beta_{\mathrm{stall}}\) bonuses for late and post-stall progress, \(c_0<0\) the zero-gain penalty, and \(\lambda_q\geq0\) the explicit query-penalty coefficient. We use \((w_{\mathrm{early}},w_{\mathrm{middle}},w_{\mathrm{late}})=(0.25,0.5,1)\), \(\beta_{\mathrm{tail}}=\beta_{\mathrm{stall}}=0.25\), \(c_0=-0.02\), and \(\lambda_q=0\). The leading term already normalizes gain by approximate attempted support \(u_t\bar g_t\); the reward is a scale-aware surrogate, not a probability.

After updating coordinate counts and appending \((a_t,\Delta L_t,u_t)\) to \(\mathcal{W}_t\), feedback updates
\begin{equation}
\begin{aligned}
Q_{t+1} &= Q_t + u_t, \\
H_{t+1} &= \begin{cases}
0, & \Delta L_t > 0,\\
H_t + 1, & \Delta L_t = 0,
\end{cases} \\
\widehat N_{t+1} &= \min\bigl(J_t, \max(1, L_{t+1})\bigr),
\end{aligned}
\label{eq:feedback-state-update}
\end{equation}
where \(Q_{t+1}\) is the updated query count, \(H_{t+1}\) resets after positive gain and otherwise increments, and \(\widehat N_{t+1}\) carries \(J_t\) clipped to the nonzero range supported by \(L_{t+1}\). The encoder then recomputes \(\mathbf{s}_{t+1}\) and valid-action mask \(\boldsymbol\mu_{t+1}\), producing
\begin{equation}
\tau_t=(\mathbf{s}_t,a_t,r_t,\mathbf{s}_{t+1},d_t,\boldsymbol\mu_{t+1}),
\label{eq:feedback-transition}
\end{equation}
where \(\tau_t\) contains current state \(\mathbf{s}_t\), action \(a_t\), reward \(r_t\), next state \(\mathbf{s}_{t+1}\), terminal flag \(d_t\), and next mask \(\boldsymbol\mu_{t+1}\). We set \(d_t=1\) at \(Q_{t+1}\geq B\), \(L_{t+1}=0\), or the image boundary, and zero otherwise. After warm-up, stored transitions update DDQN. The shaped reward is only an online signal: its necessity is not claimed, and useful state-conditioned learning is assessed against frozen, untrained, random, fixed, state-blind, and state-shuffled controls. Algorithm~\ref{alg:rlgpa} summarizes the loop.

\begin{algorithm}[tp]
\small
\caption{RLGPA continual-online support search}
\label{alg:rlgpa}
\begin{algorithmic}[1]
\Require Hard-label oracle $f$, original $\mathbf{x}$, candidate start $\mathbf{z}_0$, budget $B$, shared planner $\Pi$
\State verify $\mathcal{A}(\mathbf{z}_0)=1$; set $\mathbf{z}\gets\mathbf{z}_0$, $Q\gets1$; reset image-local state
\While{$Q<B$ and $L(\mathbf{z},\mathbf{x})>0$}
    \State encode $(\mathbf{s}_t,\boldsymbol\mu_t)$; choose $a_t=(p_t,\kappa_t)$ by masked $\epsilon$-greedy $\Pi$
    \State compute $(g_t,J_t)$; partition and order $\Omega_t$ using $p_t$
    \For{$j=1,\ldots,\min(q_c,J_t,B-Q)$}
        \State query $\widetilde{\mathbf{z}}\gets\mathcal{R}_{G_{t,j}}(\mathbf{z};\mathbf{x})$; commit iff feasible; update memory and $Q$
    \EndFor
    \State compute feedback; encode next state; store $\tau_t$; update DDQN after warm-up
\EndWhile
\State \Return $\mathbf{z}$ and the updated shared planner
\end{algorithmic}
\end{algorithm}

\section{Experiments and Evaluation}
\subsection{Experimental Setup}
\paragraph{Datasets and target models}
We evaluate CIFAR-10 with ResNet18, and ImageNet-1k with ResNet50 and ViT. The cross-method evaluation uses 100 CIFAR-10 images and 200 ImageNet images per attack mode. The query budgets of Targeted and untargeted attacks are 2,000 and 500 on CIFAR-10, and 20,000 and 5,000 on ImageNet, respectively.

\begin{table*}[tp]
        \setlength{\tabcolsep}{8pt}
	\centering
   \caption{Median sparsity and ASR at different query budgets. A comprehensive comparison among different attacks ($\sigma$-zero, PGD$_0$, Pointwise, SparseEvo, and RLGPA) on small and large scale balance datasets.  The two white-box rows are reported at their converged distortion for reference only: their optimization cost is not a sequence of victim queries, so they are excluded from every query-efficiency comparison.}

    	\begin{tabular}{c|c|c|cc|c|cc}
    		\hline
    		\multirow{2}{*}{Setting}
            &\multirow{2}{*}{Query}
            &\multirow{2}{*}{Methods}
            &\multicolumn{2}{c|}{CIFAR10 (ResNet18)}
            &\multirow{2}{*}{Query}
            &\multicolumn{2}{c}{ImageNet (ResNet50)}\\   
            \cline{4-5} \cline{7-8}
            &&&Median&ASR@0.05&&Median&ASR@0.01\\           
      	    \cline{1-8}
            \multirow{8}{*}{Untargeted}
            &&$\sigma$-zero&0.0013&100\%&&0.0001&100\%\\
            &&PGD$_0$&0.0068&100\%&&0.0005&100\%\\
            \cline{2-8}
            &\multirow{3}{*}{200}
            &Pointwise&0.0059&91.0\%&\multirow{3}{*}{2000}&0.0024&68.0\%\\
            &&SparseEvo&0.0215&85.0\%&&0.0026&87.0\%\\
            &&RLGPA&\textbf{0.0033}&\textbf{93.0\%}&&\textbf{0.0004}&\textbf{96.0\%}\\
            \cline{2-8}
            &\multirow{3}{*}{500}
            &Pointwise&0.0059&96.0\%&\multirow{3}{*}{5000}&0.0016&80.0\%\\
            &&SparseEvo&0.0215&86.0\%&&0.0023&87.5\%\\
            &&RLGPA&\textbf{0.0026}&\textbf{100.0\%}&&\textbf{0.0004}&\textbf{99.5\%}\\         
    		\hline	
            
            \multirow{8}{*}{Targeted}
            &&$\sigma$-zero&0.0036&100\%&&0.0009&100\%\\         
            &&PGD$_0$&0.0156&100\%&&0.0037&97.0\%\\       
            \cline{2-8} 
            &\multirow{3}{*}{1000}
            &Pointwise&0.4224&0.0\%&\multirow{3}{*}{10000}&0.7464&0.0\%\\   
            &&SparseEvo&0.0210&90.0\%&&0.0275&20.0\%\\    
            &&RLGPA&0.0869&24.0\%&&\textbf{0.0116}&\textbf{45.5\%}\\  
            \cline{2-8}
            &\multirow{3}{*}{2000}
            &Pointwise&0.2236&11.0\%&\multirow{3}{*}{20000}&0.6564&0.0\%\\
            &&SparseEvo&0.0181&95.0\%&&0.0070&71.5\%\\
            &&RLGPA&\textbf{0.0085}&\textbf{99.0\%}&&\textbf{0.0034}&\textbf{88.5\%}\\
    		\hline	
    	\end{tabular}
\label{tab:main-results}
	\vspace{-0.05cm}	
\end{table*}

\paragraph{Baselines}
Pointwise and SparseEvo are the numerical external baselines because they share RLGPA's decision-only oracle and sparse-support objective. Score-based sparse attacks are excluded from numerical query comparison: a score query returns substantially richer information than a top-1-label query, so treating both as an equal unit would not constitute an oracle-matched efficiency comparison. PGD$_0$ and $\sigma$-zero are included only as white-box ASR--Sparsity references. The untargeted $\sigma$-zero curves follow its published threat model; for completeness, the targeted panels use our target-class-margin extension of the same gradient-based optimizer and are not presented as an official targeted $\sigma$-zero result. Neither white-box method appears in Query--Sparsity curves or query-efficiency rankings. For a fair comparison, all numerically compared decision-based attacks use the same model preprocessing and success denominator within a setting.

\paragraph{Metrics}
On the one hand, we evaluate query efficiency using Query–Sparsity, which reports the median normalized support over hard-label queries together with the first and third quartiles; only oracle-matched decision-based methods are included in this view. On the other hand, we measure attack strength using ASR–Sparsity, defined as the percentage of evaluation images that remain adversarial after reaching a given sparsity threshold. White-box references appear only in the ASR–Sparsity view and are visually separated from black-box attacks. In addition, we report the median, mean, and 90th-percentile $L_0$ of the final perturbations. For paired ablations, we use identical manifests and starting points across configurations, and analyze the results with win/tie/loss counts and Wilcoxon signed-rank tests.

\paragraph{RLGPA configuration}
The seven-action relative planner uses $\rho=0.002$, query chunks of 32, discount $\gamma=0.8$, initial/minimum exploration $0.15/0.02$, and exploration decay $0.995$. The Dueling network has two 128-unit hidden layers. Replay capacity is 20,000, batch size is 64, warmup is 128 transitions, learning rate is $10^{-3}$, the target network is synchronized every 250 updates, and one optimizer step is taken per eligible decision.

\begin{figure*}[tp]
\centering

\captionsetup{font=footnotesize}
\captionsetup[subfloat]{
    font=scriptsize,
    captionskip=0pt,
    farskip=0pt,
    nearskip=0pt
}

\setlength{\tabcolsep}{0pt}
\renewcommand{\arraystretch}{0.85}

\newcommand{\plotcell}[1]{%
    \subfloat[]{%
        \includegraphics[width=\linewidth]{#1}%
    }%
}

\begin{tabular}{@{}
>{\raggedright\arraybackslash}m{0.08\textwidth}
@{}
*{4}{>{\centering\arraybackslash}m{0.23\textwidth}@{}}
}

&
\scriptsize\bfseries\shortstack{Untargeted\\ASR--Sparsity}
&
\scriptsize\bfseries\shortstack{Targeted\\ASR--Sparsity}
&
\scriptsize\bfseries\shortstack{Untargeted\\Query--Sparsity}
&
\scriptsize\bfseries\shortstack{Targeted\\Query--Sparsity}
\\[-0.5ex]

\scriptsize\shortstack[l]{\textbf{CIFAR-10}\\ResNet18}
&
\plotcell{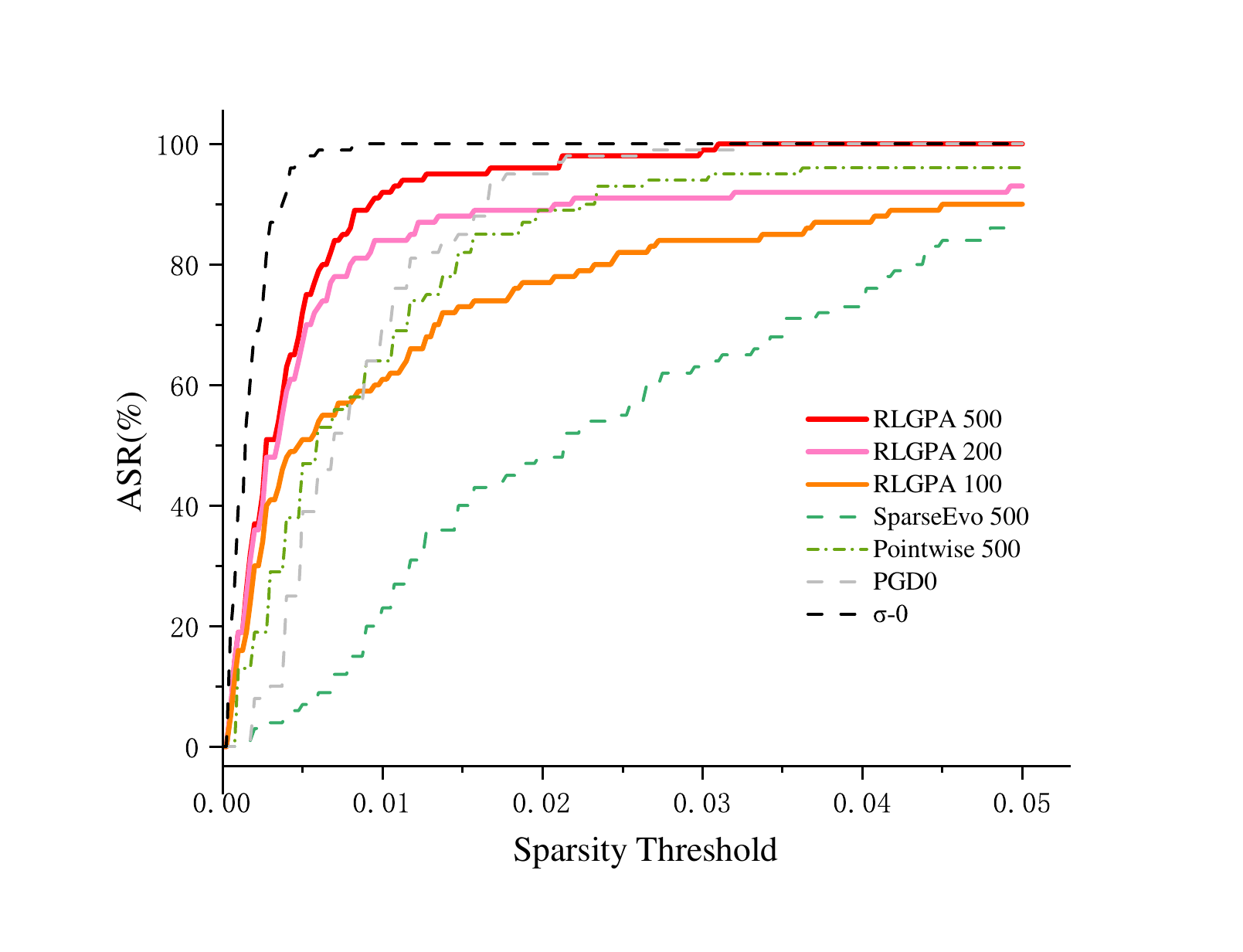}
&
\plotcell{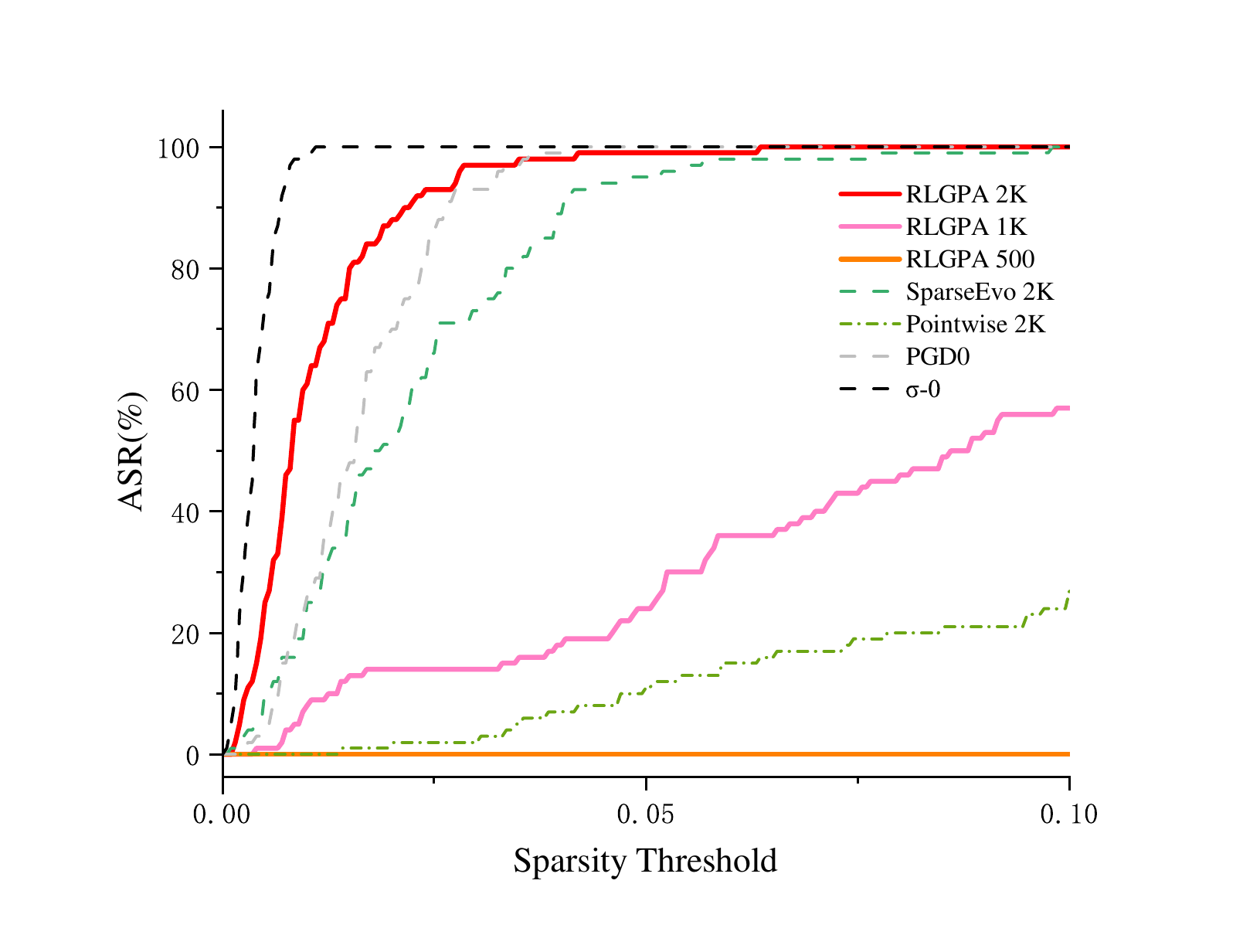}
&
\plotcell{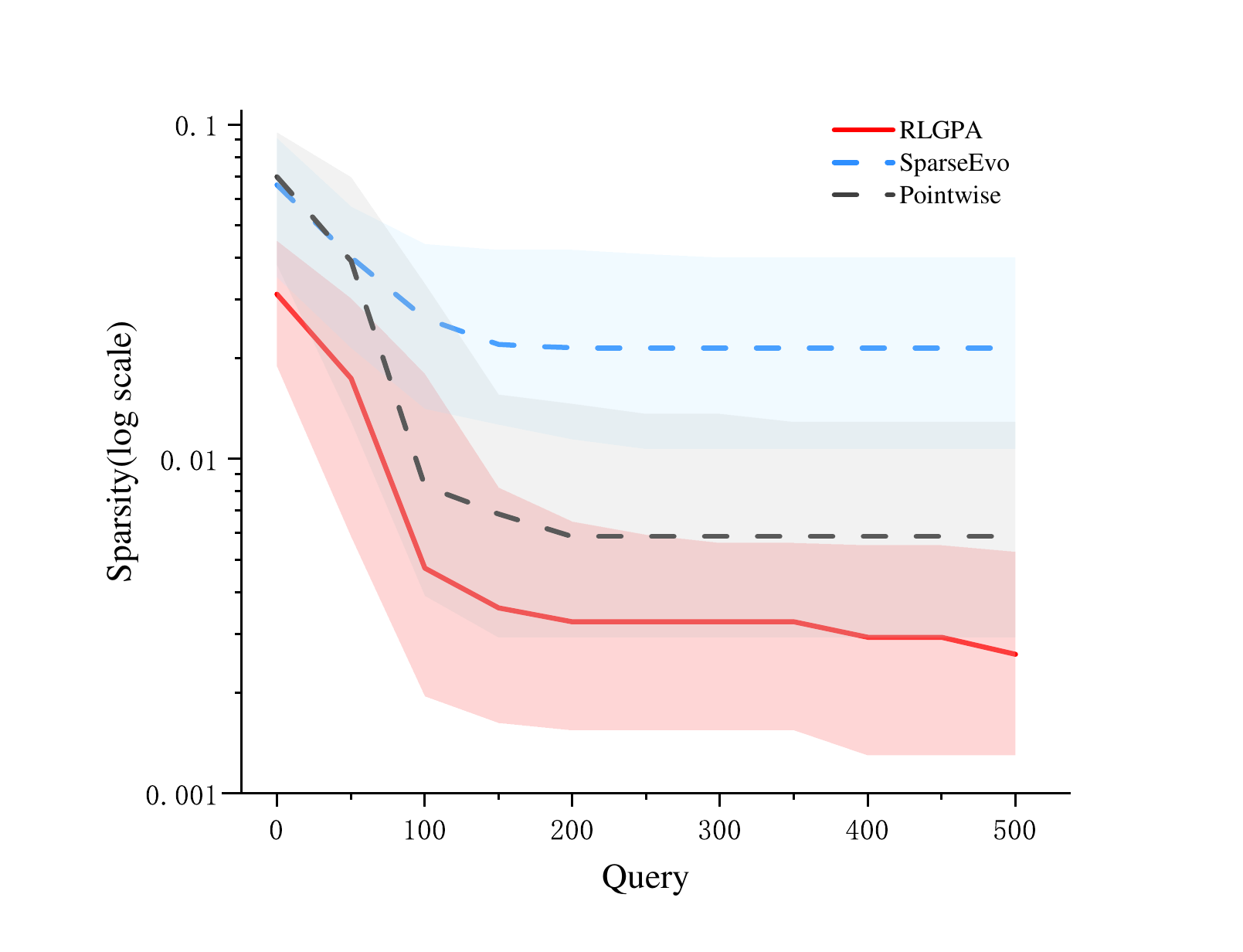}
&
\plotcell{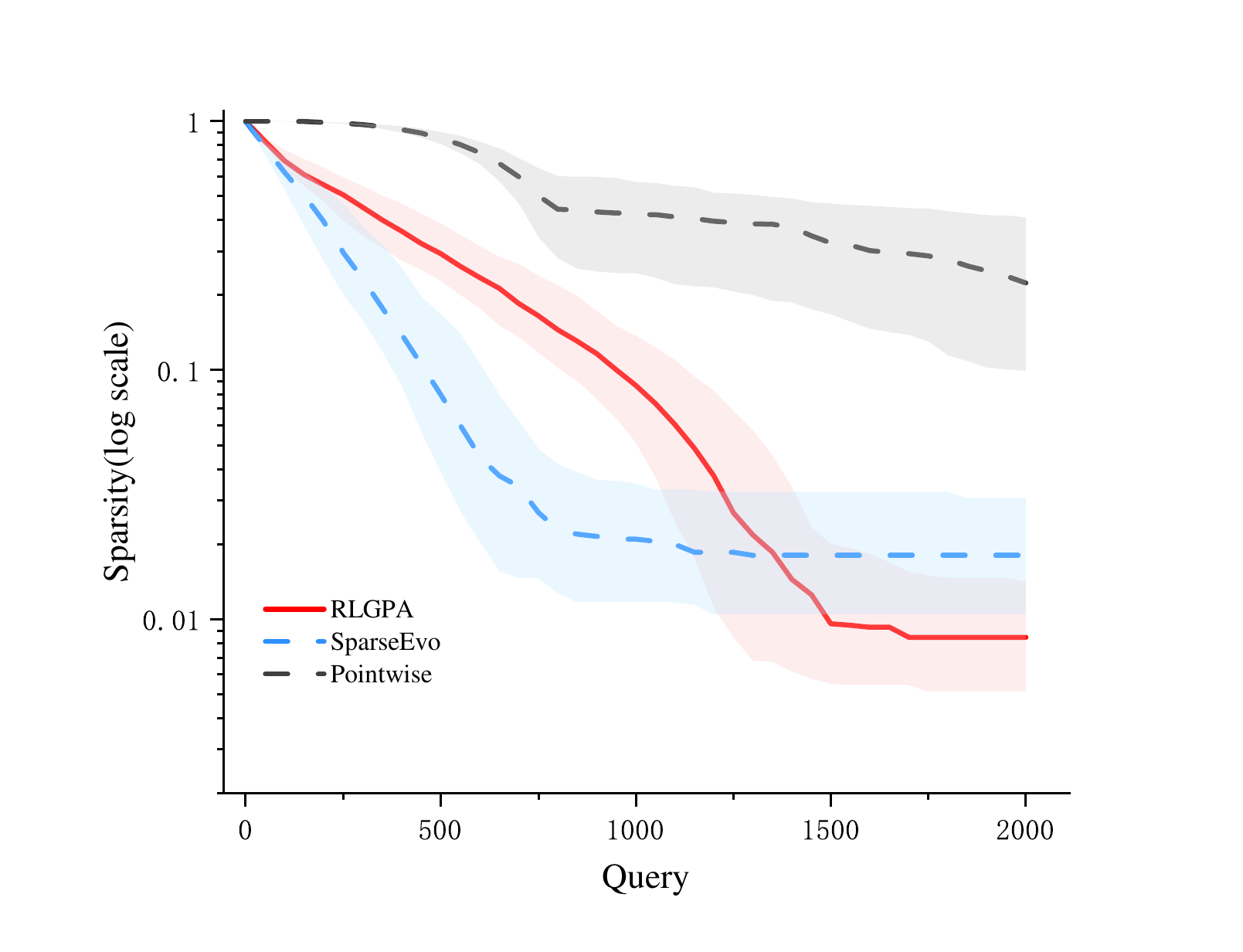}
\\[-0.5ex]

\scriptsize\shortstack[l]{\textbf{ImageNet}\\ResNet50}
&
\plotcell{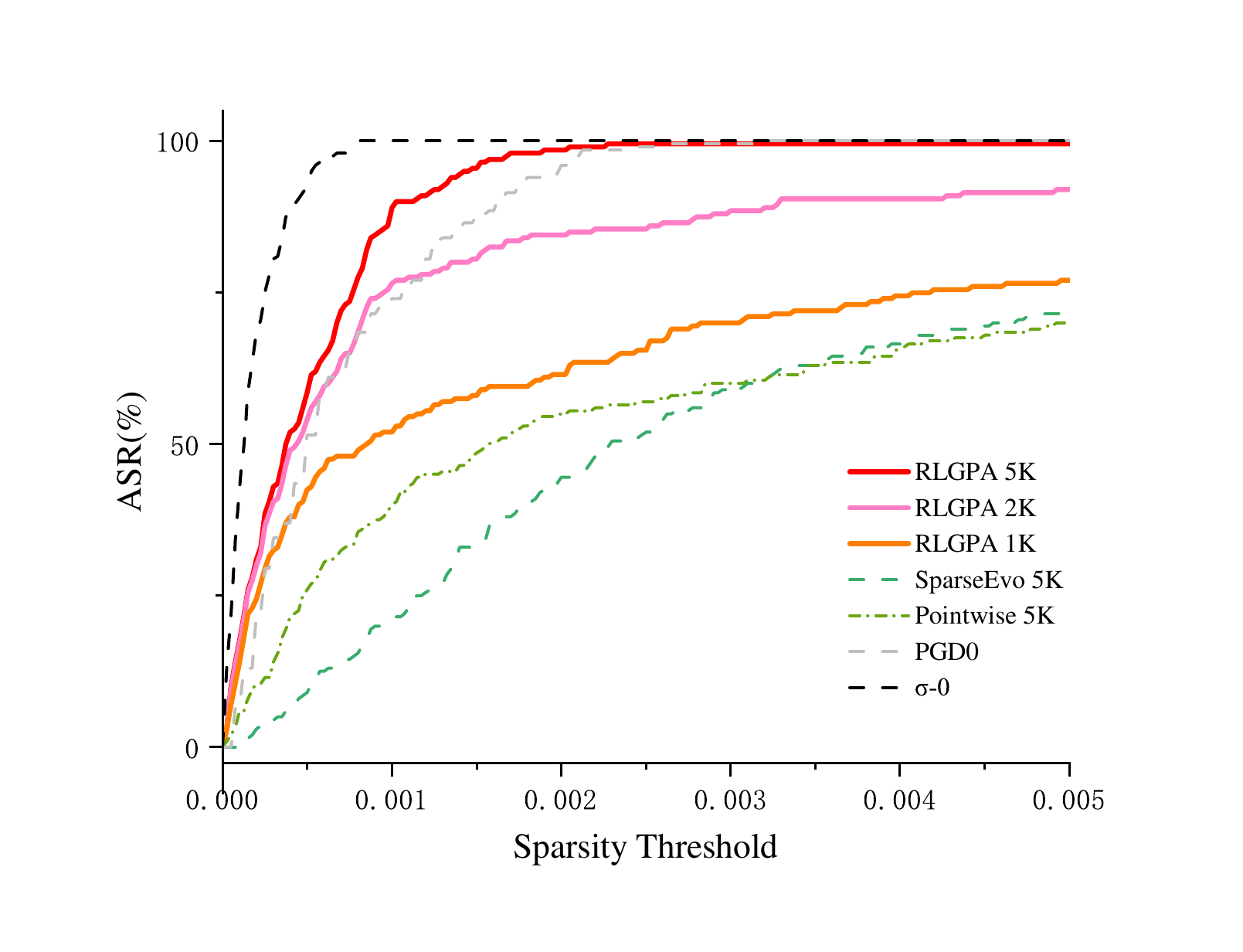}
&
\plotcell{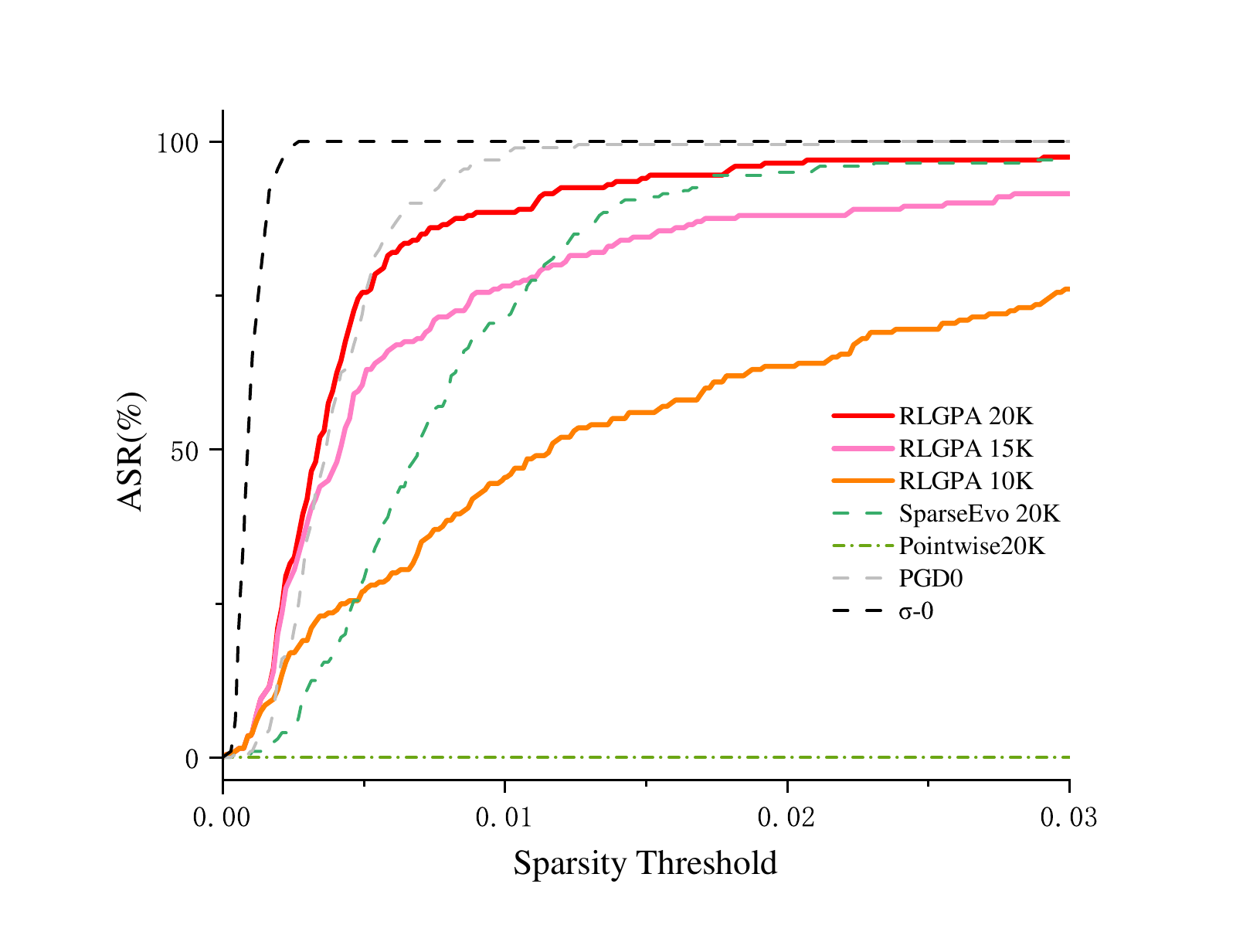}
&
\plotcell{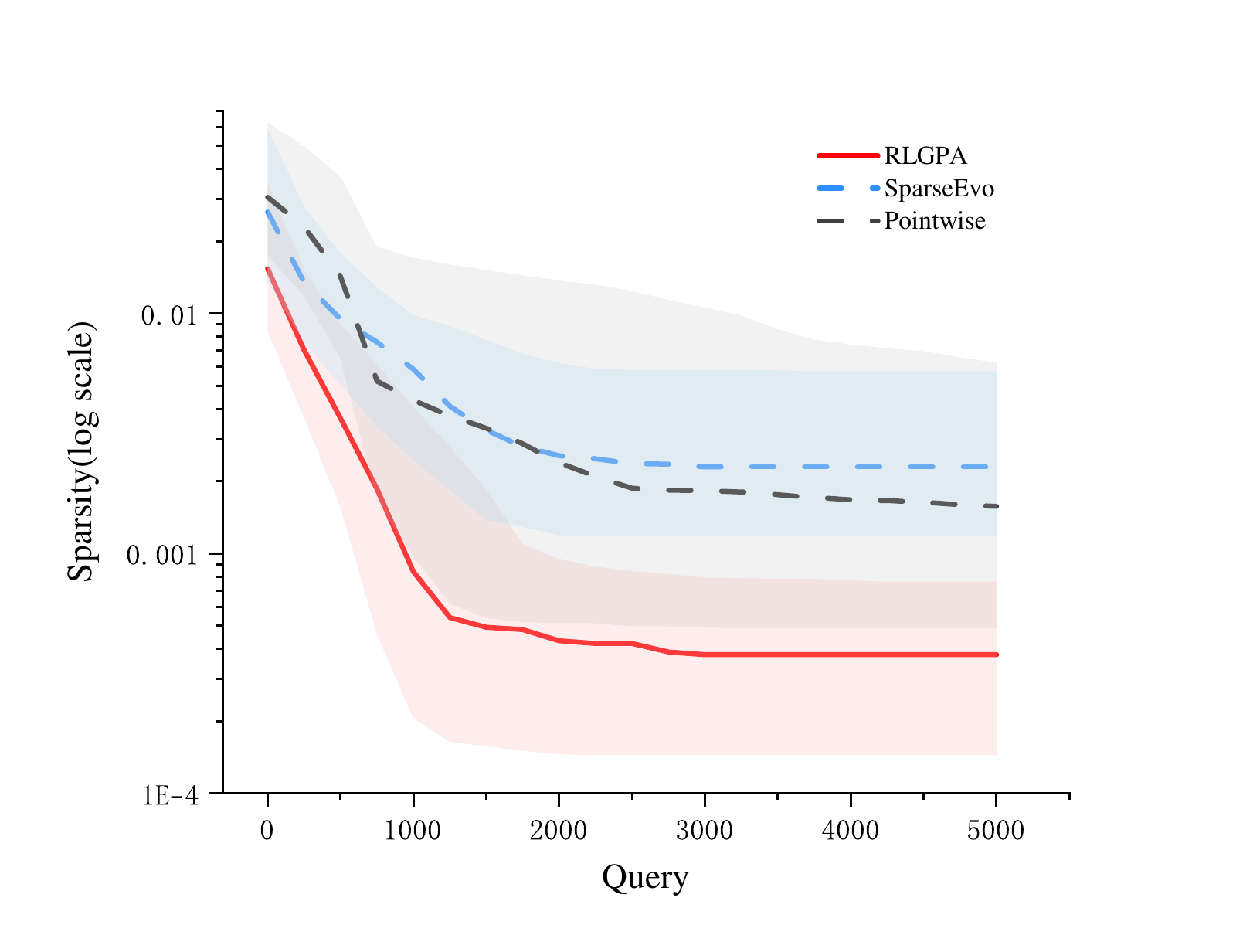}
&
\plotcell{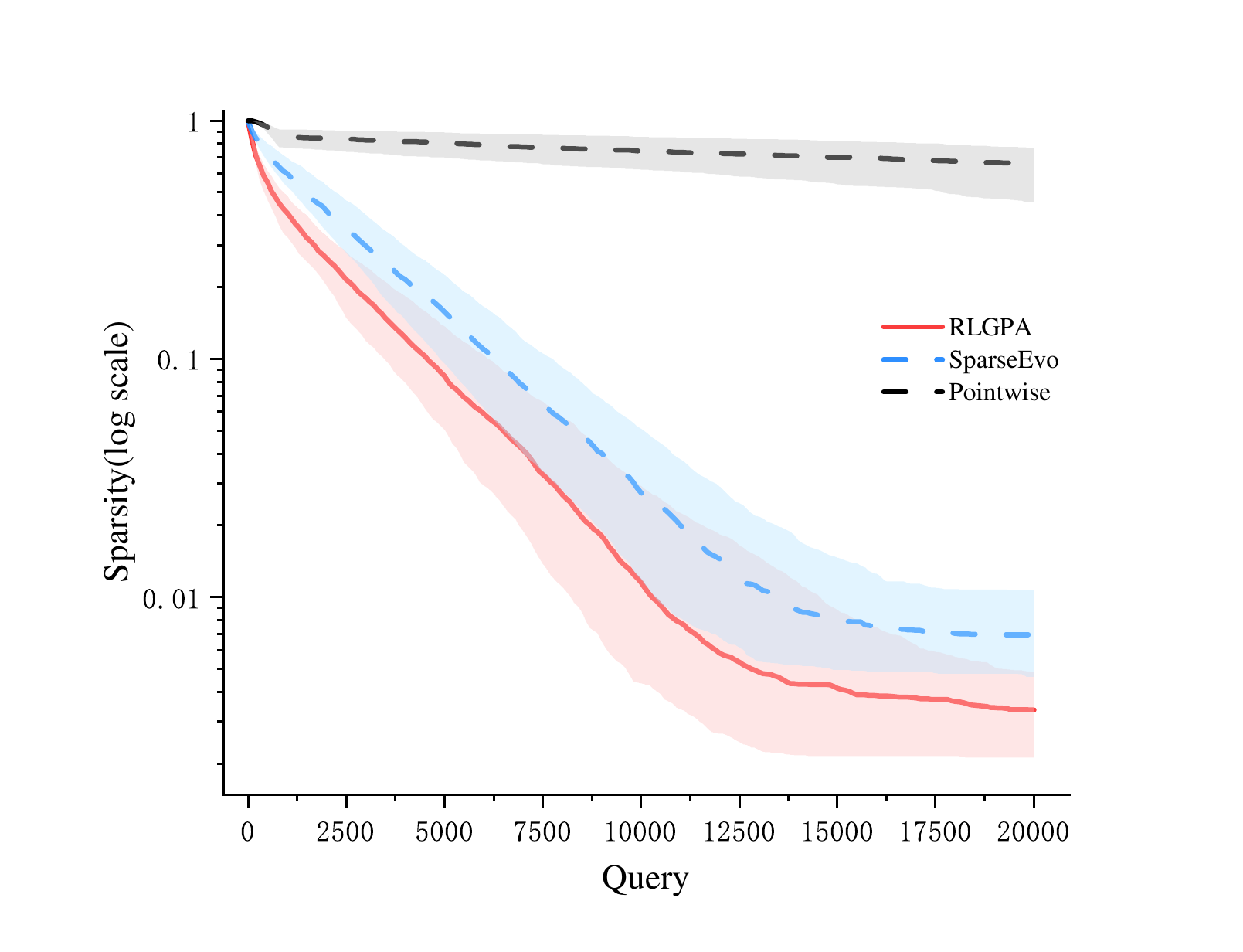}
\\[-0.5ex]

\scriptsize\shortstack[l]{\textbf{ImageNet}\\ViT}
&
\plotcell{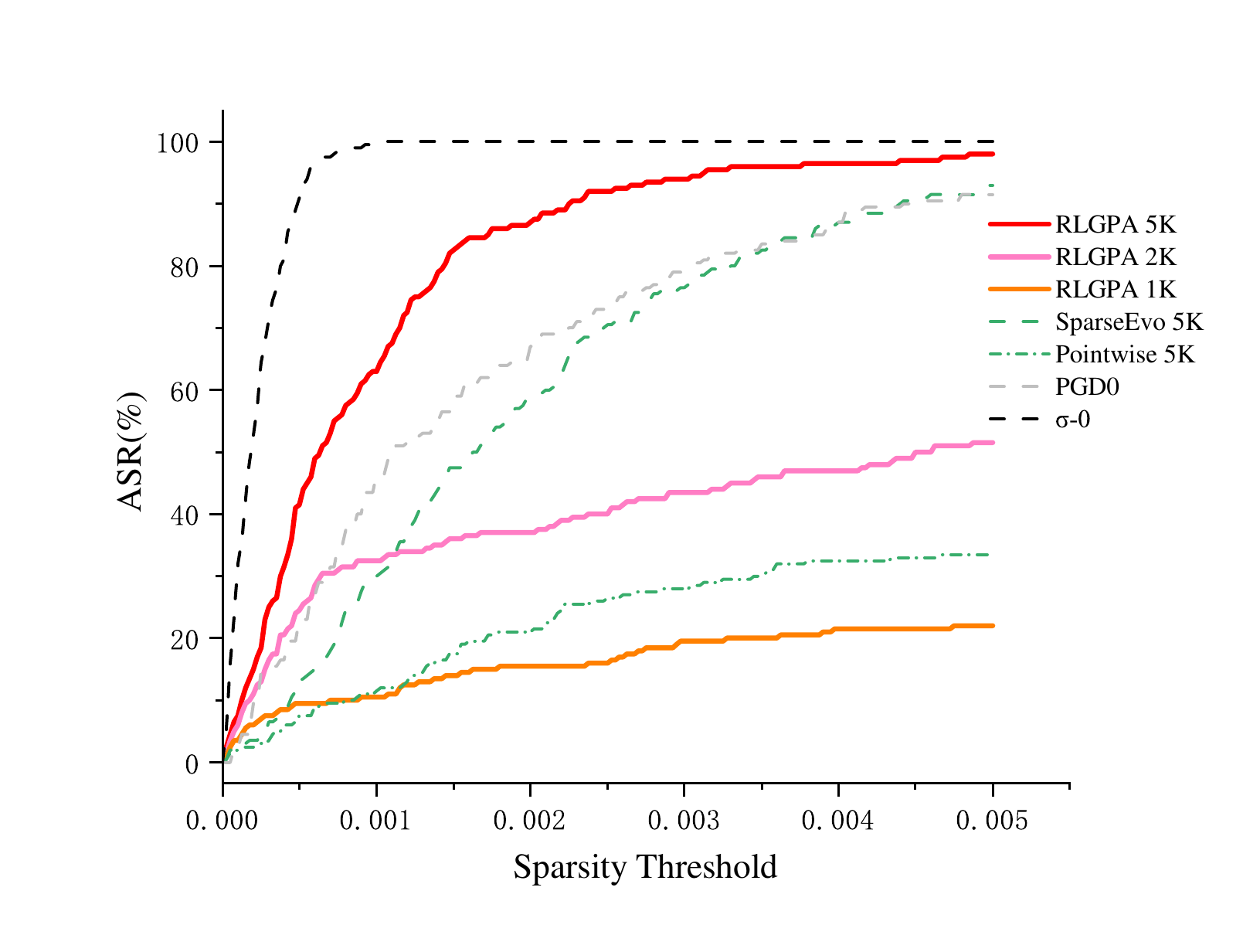}
&
\plotcell{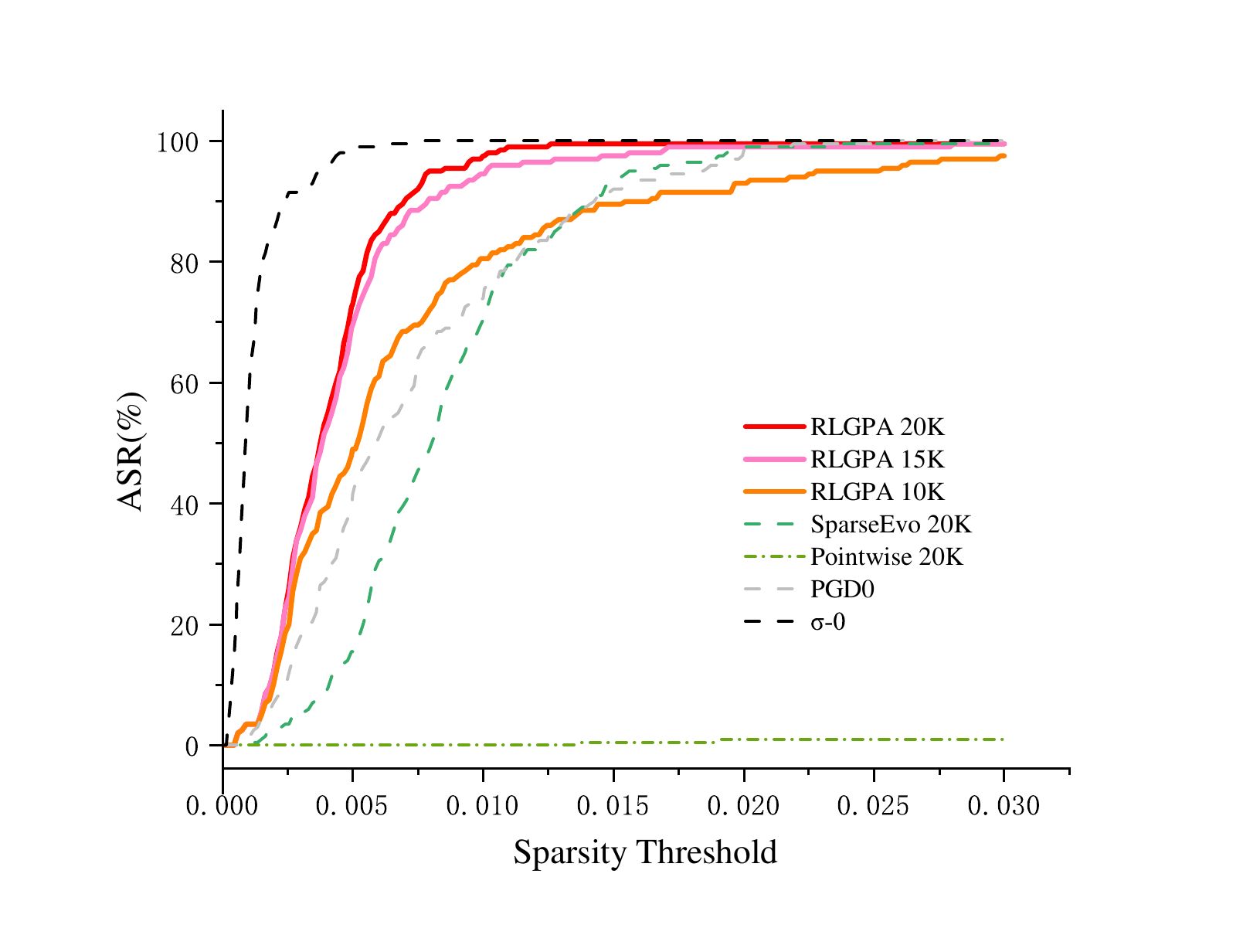}
&
\plotcell{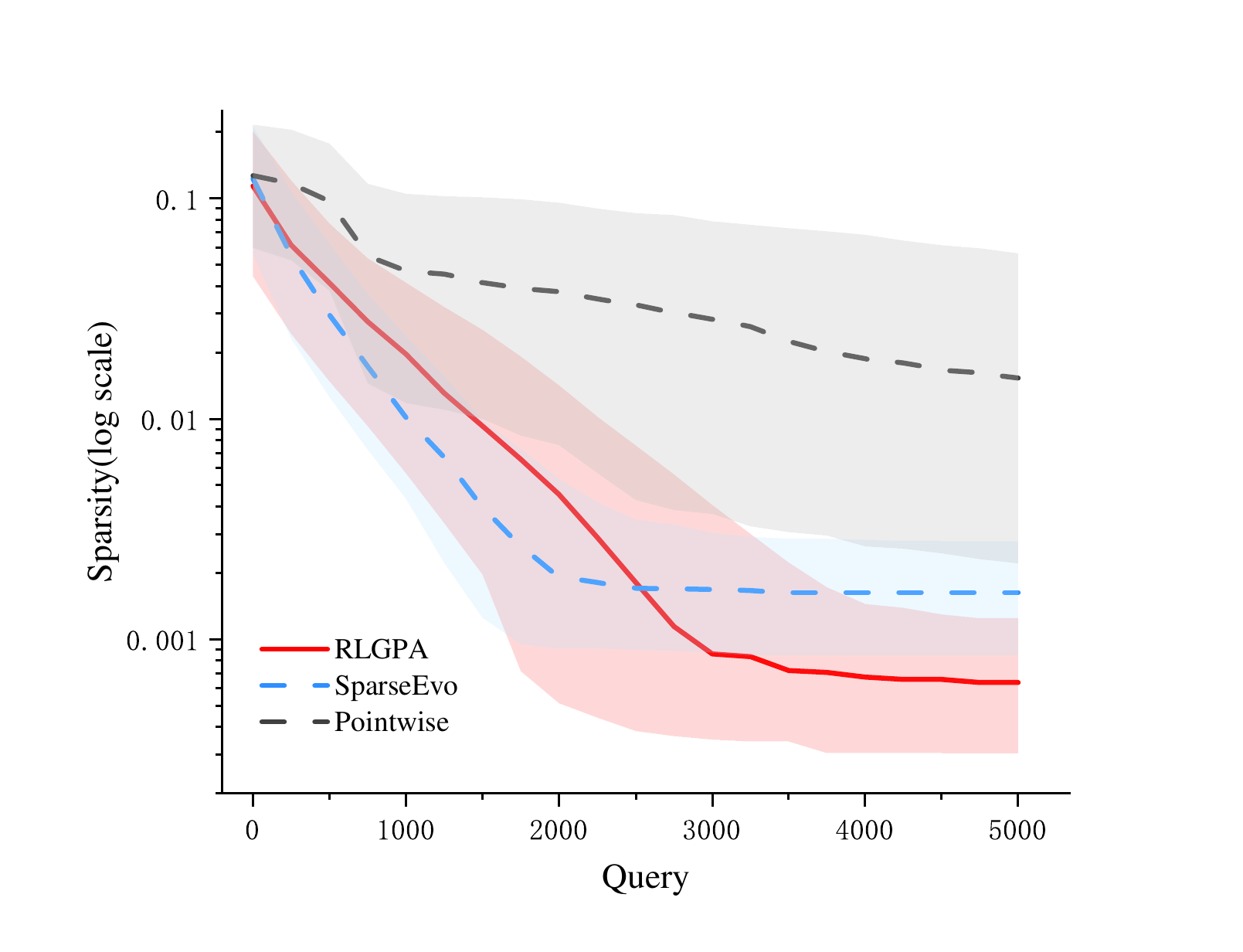}
&
\plotcell{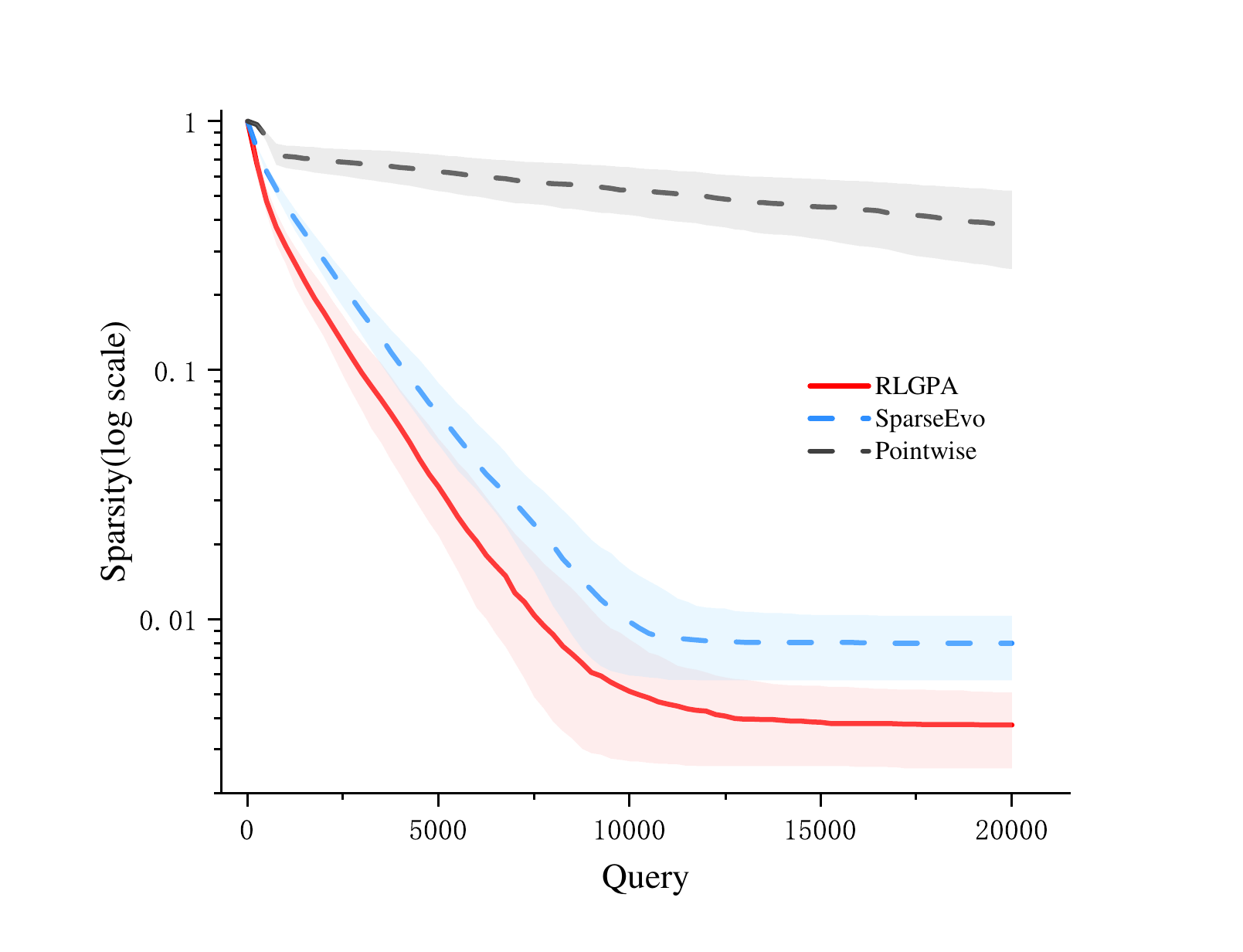}

\end{tabular}

\vspace{-1ex}

\caption{Comparison of RLGPA on CIFAR-10 (ResNet18)
and ImageNet (ResNet50 and ViT).
Rows correspond to dataset--model pairs, while columns
present untargeted/targeted ASR--Sparsity and
Query--Sparsity results.
PGD$_0$ and $\sigma$-zero are white-box references
included only in the ASR--Sparsity panels.}
\label{fig:combined-main}

\end{figure*}

\subsection{Main Comparison}
\label{Main Comparison}
Table~\ref{tab:main-results} summarizes the final-budget results under the matched top-1-label oracle. We use a strict threshold of 0.01 for CIFAR-10 targeted attacks, 0.005 for ImageNet targeted attacks, and 0.001 for untargeted attacks. Full threshold curves are given in Figure~\ref{fig:combined-main}; the conclusion does not depend on a single operating point. PGD\(_0\) and \(\sigma\)-zero are reference curves for sparse robustness only and are excluded from the query-efficiency claim.

\begin{table}[tp]
        \setlength{\tabcolsep}{3pt}
	\centering
   \caption{ASR of ResNet50 and ViT under attacks at different query budgets and sparsity thresholds. A comprehensive comparison among different attacks ($\sigma$-zero, PGD$_0$, SparseEvo, and RLGPA) on small and
large scale balanced evaluation sets from ImageNet.}

        
    	\begin{tabular}{c|c|c|cc|cc}
    		\hline
    		Setting&Query&Methods 
            & \multicolumn{2}{c|}{ResNet50}
            &\multicolumn{2}{c}{ViT}\\
            
            \hline	
            
            \multicolumn{3}{c|}{Sparsity}
            &0.001&0.005&0.001&0.005\\
            
            \hline	
            
            \multirow{8}{*}{Untargeted}
            &&$\sigma$-zero&100\%&100\%&100\%&100\%\\
            
            &&PGD$_0$&74.0\%&100.0\%&46.5\%&91.5\%\\
            
            \cline{2-7}
            
            &\multirow{2}{*}{2000}
            &SparseEvo&21.0\%&68.5\%&27.0\%&74.0\%\\
            &&RLGPA&\textbf{76.5\%}&\textbf{92.0\%}&\textbf{32.5\%}&51.5\%\\
            
            \cline{2-7}
            
            &\multirow{2}{*}{5000}
            &SparseEvo&21.5\%&72.0\%&30.0\%&93.0\%\\
            &&RLGPA&\textbf{89.0\%}&\textbf{99.5\%}&\textbf{63.0\%}&\textbf{98.0\%}\\
    		
            \hline	
            
            \multicolumn{3}{c|}{Sparsity}
            &0.001&0.005&0.001&0.005\\
            
            \hline	
            
            \multirow{8}{*}{Targeted}
            &&$\sigma$-zero&60.5\%&100\%&57.5\%&98.0\%\\
            
            &&PGD$_0$&1.0\%&73.5\%&1.5\%&41.5\%\\
            
            \cline{2-7}
            
            &\multirow{2}{*}{10000}
            &SparseEvo&0.0\%&11.5\%&0.5\%&14.0\%\\
            &&RLGPA&\textbf{3.5\%}&\textbf{27.0\%}&\textbf{3.5\%}&\textbf{49.0\%}\\
            
            \cline{2-7}
            
            &\multirow{2}{*}{20000}
            &SparseEvo&0.5\%&29.0\%&0.5\%&15.5\%\\
            &&RLGPA&\textbf{3.5\%}&\textbf{75.5\%}&\textbf{3.5\%}&\textbf{73.0\%}\\
    		
            \hline    		
    	\end{tabular}

\label{tab:victim-comparison}
	\vspace{-0.5cm}	
\end{table}


On CIFAR-10, RLGPA achieves lower median sparsity and higher ASR than Pointwise and SparseEvo at both untargeted budgets. In the targeted setting, Pointwise largely fails, SparseEvo is the strongest baseline, and RLGPA is weaker at the half budget but clearly best at the full budget (median sparsity 0.0085, ASR 99.0\%). This pattern is consistent with the online-learning characterization in Section~\ref{sec:probe}: the shared planner improves over the stream, so its advantage concentrates at larger budgets.


On ImageNet, the same ordering holds. Untargeted median sparsity remains at 0.0004 with 96.0\% and 99.5\% ASR, and targeted median sparsity improves from 0.0116 to 0.0034 with ASR rising from 45.5\% to 88.5\% as the budget grows. Notably, RLGPA attains sparsity levels comparable to the white-box methods \(\sigma\)-zero and PGD\(_0\), despite using only hard-label feedback.


Table~\ref{tab:victim-comparison} further reports ImageNet results separately for ResNet50 and ViT at sparsity 0.001 and 0.005. Under both untargeted and targeted settings, RLGPA leads at strict sparsity and full budget on both architectures. SparseEvo retains an advantage only in the lowest-budget, loose-threshold ViT cell. The transformer result confirms that the policy--executor decomposition is not tied to convolutional victims.

Figure~\ref{fig:combined-main} confirms that RLGPA leads in the low-sparsity region across all three settings and throughout the budget range. White-box references appear only in the ASR--Sparsity panels and are visually separated from the black-box curves. Overall, RLGPA dominates the decision-based field at strict sparsity and full budget on both CNN and transformer victims, conceding only the lowest-budget, loose-threshold ViT cells to SparseEvo.

\begin{figure}[tp]
\centering
\includegraphics[width=\linewidth]{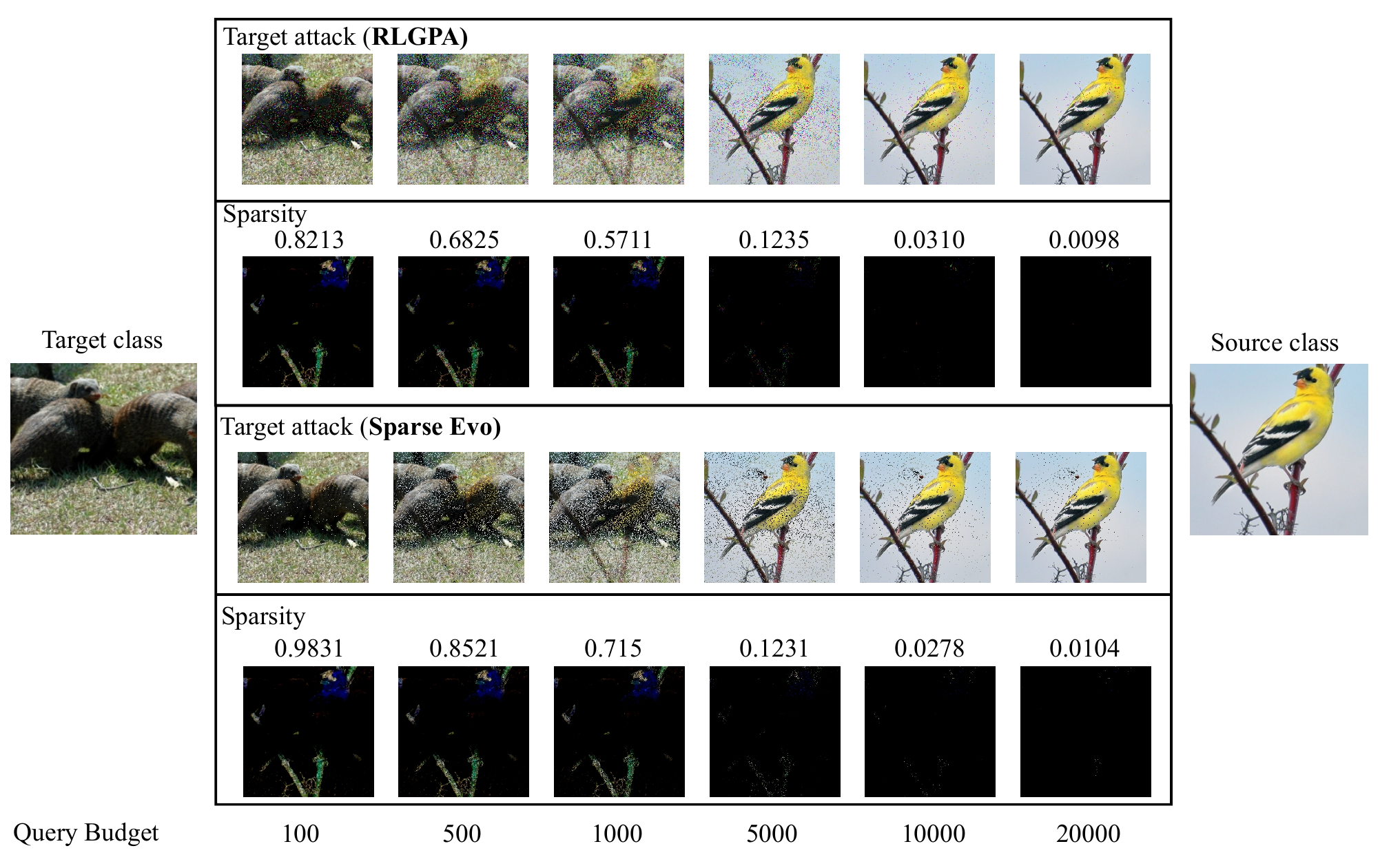}
\caption{Visual comparison of targeted attacks between RLGPA and SparseEvo on ImageNet. Query budgets: 100, 500, 1,000, 5,000, 10,000, and 20,000. For each method, the upper row shows the adversarial image and the lower row its perturbation map with normalized sparsity. Target-class reference (left) and source image (right).}
\label{fig:qualitative-comparison}
\end{figure}


\begin{figure}[tp]
\centering
\includegraphics[width=\linewidth]{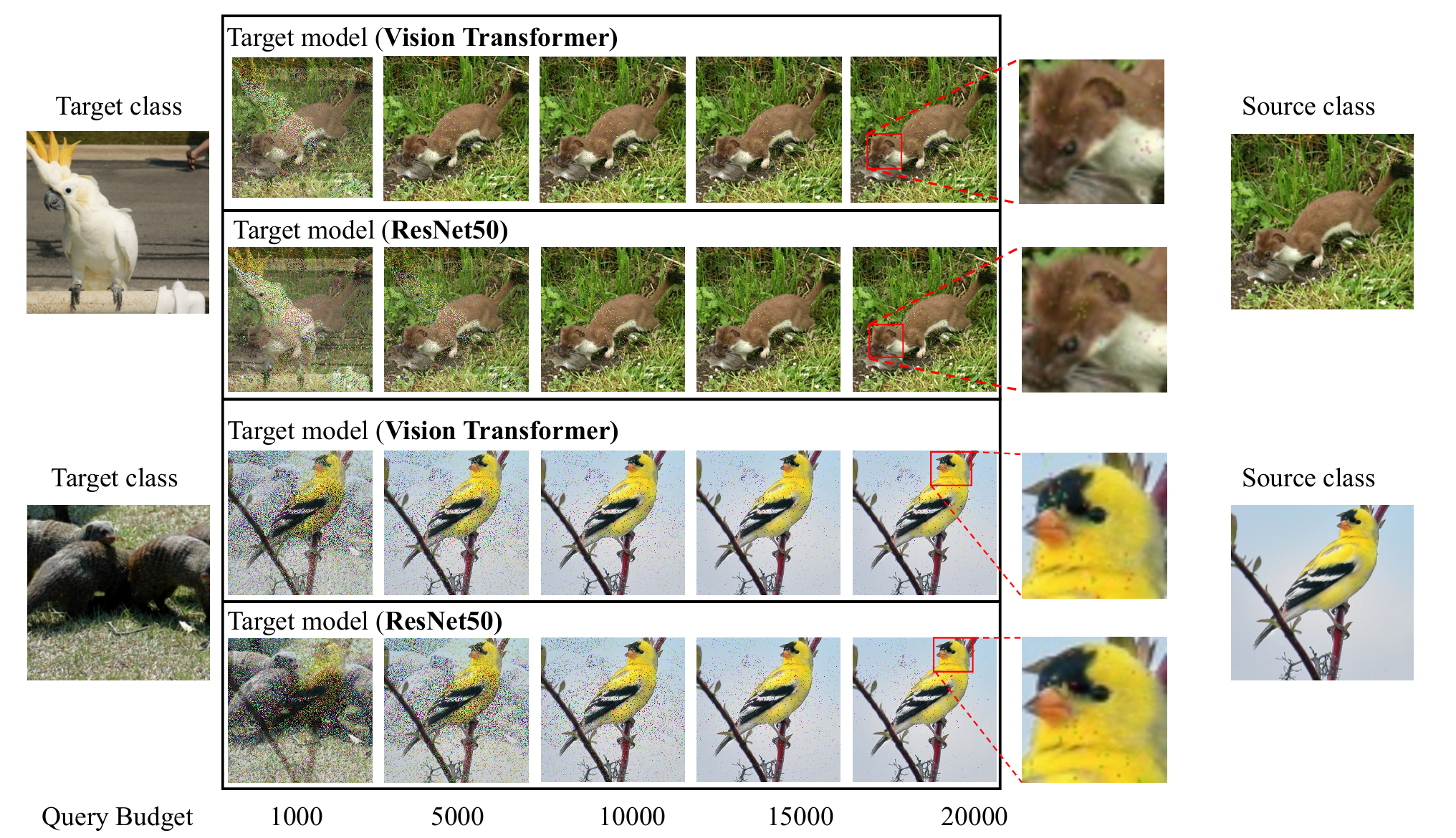}
\caption{Visual evolution of targeted adversarial examples generated by RLGPA on ViT and ResNet50 on ImageNet. Two source–target pairs are shown. Query budgets: 1,000, 5,000, 10,000, 15,000, and 20,000. Target-class reference (left), source image (right); enlarged crops highlight residual perturbations at the final budget.}
\label{fig:qualitative-architectures}
\end{figure}

\subsection{Qualitative Evolution Across Queries}
\label{sec:qualitative-visualization}

Section~\ref{Main Comparison} quantitatively demonstrate that the RLGPA attack achieves strong performance on both CNN and Transformer victim models. By contrast, Figures~\ref{fig:qualitative-comparison} and~\ref{fig:qualitative-architectures} qualitatively illustrate the attack effectiveness of different methods by showing how individual targeted adversarial examples evolve with the query budget.

Figure~\ref{fig:qualitative-comparison} shows that both attacks progressively restore the visual structure of the source image while retaining targeted feasibility. On this example, RLGPA contracts the diffuse perturbation more rapidly at the early budgets and reaches a lower final sparsity at 20,000 queries. The perturbation maps make the support contraction visible rather than relying only on the scalar sparsity values.

Figure~\ref{fig:qualitative-architectures} further illustrates the same restoration process on both a convolutional and a transformer victim. At lower budgets, the perturbation remains spatially widespread; as the query budget increases, the adversarial image progressively resembles the source image, and the residual changes become localized. The similar trends observed on ResNet50 and ViT provide a qualitative complement to the architecture-specific quantitative results in Table~\ref{tab:victim-comparison}.

\subsection{Does the Planner Learn? Fixed-Probe Evaluation}
\label{sec:probe}
The main curves measure continual-online attack performance but do not by themselves prove learning, because image difficulty can vary with stream position. We therefore construct four non-overlapping CIFAR-10 targeted splits: 100 attribution images, 120 adaptation images, 20 validation images, and 20 fixed probe images. Checkpoints are frozen and evaluated with $\epsilon=0$; parameter hashes, optimizer steps, and replay sizes are checked before and after every validation/probe run. The probe split is never used for checkpoint selection, so any improvement on it reflects transfer rather than fitting to the evaluation images.

Table~\ref{tab:probe} reports the frozen performance on the fixed, disjoint 20-image probe set across different numbers of adaptation images. No probe-time parameter or replay update is allowed. At 0 adaptation images, the median \(L_0\) is 165.0 and the ASR@0.01 is 0.20. As the number of adaptation images increases, the median \(L_0\) decreases steadily---reaching 23.0 at 40 adaptation images and stabilizing around 21.5 at 120 adaptation images---while ASR@0.01 rises from 0.20 to 0.65. These results show that the planner improves on the probe set as it accumulates attack experience, without any test-time updates.

\begin{table}[t]
\centering
\caption{Frozen performance on a fixed, disjoint 20-image probe set. No probe-time parameter or replay update is allowed.  Median, Mean, and P90 are for \(L_0\).}
\label{tab:probe}
\setlength{\tabcolsep}{4pt}
\begin{tabular}{rrrrr}
\toprule
Adaptation images & ASR@.01 & Median & Mean & P90 \\
\midrule
0 & .20 & 165.0 & 136.20 & 226.2 \\
20 & .45 & 41.5 & 40.65 & 70.6 \\
40 & .60 & 23.0 & 28.20 & 57.4 \\
60 & .50 & 25.5 & 27.75 & 55.7 \\
80 & .55 & 21.5 & 27.55 & 52.0 \\
100 & .60 & \textbf{21.0} & 28.35 & 54.1 \\
120 & \textbf{.65} & 21.5 & \textbf{25.15} & \textbf{50.2} \\
\bottomrule
\end{tabular}
\end{table}

To isolate this transfer effect, we select checkpoint 40 using the validation split only and compare it against checkpoint 0 on the 20 fixed probe images. All 20 probe images improve, the median paired reduction is 131 coordinates, and the Wilcoxon test gives $p = 1.91\times10^{-6}$. This improvement is accompanied by a stable evaluation state: across all 14 frozen validation/probe runs, parameter SHA256, optimizer-step count, and replay size remain unchanged. Together, these results provide the clearest evidence that prior attack experience produces reusable policy parameters rather than merely improving later images through an easier stream order.

\subsection{Matched Planner Attribution}

\begin{table}[t]
\centering
\caption{Matched planner attribution on targeted CIFAR-10. The final column is the paired Wilcoxon test against Full DQN.  Median, Mean, and P90 are for \(L_0\).}
\label{tab:planner-attribution}
\setlength{\tabcolsep}{2pt}
\begin{tabular}{lrrrrr}
\toprule
Planner/control & ASR@.01 & Median & Mean & P90 & $p$ \\
\midrule
Full shared-online DQN & \textbf{.59} & \textbf{24.5} & 35.56 & 67.0 & -- \\
Base Memory only & .57 & 25.5 & \textbf{33.65} & 63.1 & .8535 \\
Deterministic schedule & .50 & 30.5 & 33.45 & \textbf{57.1} & .1117 \\
Uniform random & .46 & 35.0 & 38.04 & 68.5 & \textbf{.0028} \\
Train-disabled DQN & .21 & 122.5 & 112.32 & 199.1 & $<10^{-15}$ \\
Per-image-reset DQN & .21 & 122.5 & 112.32 & 199.1 & $<10^{-15}$ \\
\bottomrule
\end{tabular}
\end{table}

Table~\ref{tab:planner-attribution} compares planners using the same 100-image targeted manifest, starting points, executor, and 2,000-query budget. Full shared-online DQN significantly outperforms uniform random, train-disabled, and per-image-reset controls. It is not significantly better than the strongest fixed Base Memory policy or the deterministic schedule on this set. The fixed-probe result above is therefore the primary learning evidence, while this experiment identifies the relevant baselines and limits the claim.

To further test whether the learned policy is genuinely necessary, we also examine how much of the observed performance can be attributed to the network's initialization rather than to adaptation. Five frozen-untrained network seeds are highly unstable, with aggregate median \(L_0 = 217.5 \pm 169.3\); the best seed reaches median 29.0. Thus a favorable random output bias can occasionally imitate a useful fixed action preference, but it is not a reliable substitute for adaptation.

\subsection{State Conditioning and Reward}
\label{sec:state-ablation}

Table~\ref{tab:state-ablation} removes groups of state features while retaining the same action space and executor. Full state significantly outperforms state-blind inference with 66/9/25 wins/ties/losses and $p=3.76\times10^{-9}$. It also outperforms each of five state-shuffled seeds at $p<0.05$. The final relative planner therefore uses state alignment rather than learning only a global action prior.

\begin{table}[t]
\centering
\small
\setlength{\tabcolsep}{2pt}   
\caption{State ablation for the final relative planner. Median, Mean, and P90 are for \(L_0\).}
\label{tab:state-ablation}
\begin{tabular}{lrrrr}
\toprule
State variant & \makecell{ASR\\@.01} & Median & Mean & P90 \\
\midrule
Full state & \textbf{.59} & \textbf{24.5} & \textbf{35.56} & \textbf{67.0} \\
No action history/effectiveness & .54 & 27.0 & 37.77 & 71.1 \\
No recent progress/stall & .54 & 29.0 & 38.64 & 73.7 \\
No phase/budget & .50 & 30.5 & 36.41 & 71.0 \\
Minimal legal mask & .48 & 31.5 & 46.67 & 90.2 \\
State-shuffled, five-seed mean & .482 & 32.8 & 45.35 & 89.16 \\
State-blind & .44 & 44.5 & 82.01 & 203.4 \\
\bottomrule
\end{tabular}
\end{table}

The reward ablation in Table~\ref{tab:reward-ablation} presents a more nuanced picture. Removing individual phase or tail terms degrades performance, yet the full shaped reward is statistically indistinguishable from a simple support-gain-per-query reward (\(p = .8392\)). We therefore do not claim that the shaped reward is intrinsically superior.

\begin{table}[t]
\centering
\small
\setlength{\tabcolsep}{5pt}
\caption{Reward ablation on targeted CIFAR-10.  Median, Mean, and P90 are for \(L_0\).}
\label{tab:reward-ablation}

\begin{tabular}{lrrrr}
\toprule
Reward &\makecell{ ASR\\@.01} & Median & Mean & P90 \\
\midrule
Support gain/query & .56 & \textbf{24.0} & \textbf{34.46} & 68.0 \\
Full shaped reward & \textbf{.59} & 24.5 & 35.56 & \textbf{67.0} \\
No stall bonus & .52 & 29.0 & 39.00 & 71.1 \\
No tail/stall bonuses & .50 & 30.0 & 39.42 & 74.2 \\
No phase weights & .49 & 31.0 & 40.29 & 72.2 \\
No tail bonus & .47 & 31.0 & 44.00 & 71.3 \\
\bottomrule
\end{tabular}
\end{table}

\subsection{Granularity and Executor Composition}
\label{sec:components}

Table~\ref{tab:component-ablation} separates scale selection from executor choice. Dynamic repartition is indispensable: fixing the partition increases median $L_0$ from 24.5 to 128. Removing Memory substantially worsens the mean and tail, whereas Memory-only remains competitive in the median. Learned scale selection is not significantly better than fixed base scale or random scale. Thus the main learned contribution is state-conditioned executor scheduling, with granularity planning providing a useful shared action representation but not yet an independently significant gain.

\begin{table}[t]
\centering
\caption{Relative granularity and executor ablation. Median, Mean, and P90 are for \(L_0\).}
\label{tab:component-ablation}
\setlength{\tabcolsep}{3.5pt}
\begin{tabular}{lrrrr}
\toprule
Variant & ASR@.01 & Median & Mean & P90 \\
\midrule
Erase ratio .01 & \textbf{.63} & \textbf{23.0} & \textbf{29.83} & \textbf{58.2} \\
Full seven-action, ratio .002 & .59 & 24.5 & 35.56 & 67.0 \\
Erase ratio .005 & .59 & 25.0 & 30.70 & 63.2 \\
Policy DQN, fixed base scale & .62 & 25.5 & 34.66 & 60.1 \\
Policy DQN, random scale & .54 & 26.5 & 36.23 & 66.1 \\
Memory-only & .52 & 28.0 & 38.18 & 68.4 \\
No Memory actions & .51 & 30.0 & 79.36 & 184.5 \\
Erase ratio .001 & .29 & 59.5 & 119.16 & 313.1 \\
Fixed partition & .21 & 128.0 & 336.59 & 1115.0 \\
Primary-only & .36 & 173.0 & 235.28 & 625.1 \\
Safe-only & .03 & 301.0 & 343.95 & 569.1 \\
\bottomrule
\end{tabular}
\end{table}

The apparent advantage of ratio .01 over .002 on CIFAR-10 is not significant in the paired comparison ($p=.0861$) and has not been repeated across CIFAR-10 training seeds. We therefore retain .002 as the common CIFAR-10/ImageNet default rather than selecting the most favorable row after observing the evaluation set. The independent three-seed ImageNet sensitivity study below further tests this fixed choice without retuning the main experiments.

\subsection{Lifecycle of Online Experience}
\label{sec:lifecycle}

Table~\ref{tab:lifecycle} studies what is inherited across images. A planner adapted on 120 images and then frozen reaches median $L_0=26$ without test-time updates, close to continual online learning and far better than per-image reset. Retaining weights while clearing replay performs poorly, but it also receives only 147 optimizer updates because most single images do not reach the 128-transition warmup. This comparison therefore establishes the importance of the complete cross-image learning state but does not isolate replay-distribution effects from update count.

\begin{table}[t]
\centering
\caption{Cross-image learning lifecycle on the matched targeted attribution set. Median, Mean, and P90 are for \(L_0\).}
\label{tab:lifecycle}
\begin{adjustbox}{max width=\textwidth}
\setlength{\tabcolsep}{2pt}
\begin{tabular}{lrrrrr}
\toprule
Lifecycle & \makecell{ASR\\@.01} & Median & Mean & P90 & Updates \\
\midrule
Keep weights + replay & \textbf{.59} & \textbf{24.5} & 35.56 & 67.0 & 7936 \\
Preadapt, frozen inference & .58 & 26.0 & \textbf{29.60} & \textbf{60.0} & 0 \\
\makecell{Reset weights + \\replay per image} & .21 & 122.5 & 112.32 & 199.1 & 107 \\
\makecell{Keep weights, clear \\replay per image} & .20 & 126.5 & 173.73 & 361.6 & 147 \\
\bottomrule
\end{tabular}
\end{adjustbox}
\end{table}

\subsection{Mechanistic Analysis}
\label{sec:mechanism}

The learned behavior agrees with the intended stage structure. Table~\ref{tab:phase-mechanism} reports the most productive action in each phase. Coarse Primary dominates early and middle support contraction, while Base Memory is markedly more effective in the late phase. In that late phase, Base Memory obtains support gain/query 0.459, compared with 0.038 for Base Primary and 0.096 for Base Safe. Combined with the state-shuffle result, this provides a mechanism-level explanation for why state-conditioned switching improves the tail.

\begin{table}[tp]
\centering
\caption{Most productive action by support-search phase.}
\label{tab:phase-mechanism}
\setlength{\tabcolsep}{2pt}
\begin{tabular}{llrl}
\toprule
Phase & Action & \makecell{Support gain\\/query} & Interpretation \\
\midrule
Early & Coarse Primary & 6.970 & aggressive initial contraction \\
Middle & Coarse Primary & 1.884 & continued coarse removal \\
Late & Base Memory & 0.459 & history-guided tail refinement \\
\bottomrule
\end{tabular}
\end{table}

Every reported matched ablation records zero illegal commits. The feasibility invariant is therefore preserved across planner, state, reward, granularity, and lifecycle variants.

\subsection{Hyperparameter Sensitivity}
\label{sec:hparam-sensitivity}
\begin{figure}[tp]
\centering
\subfloat[Relative erase ratio \(\rho\).]{\includegraphics[width=0.49\linewidth]{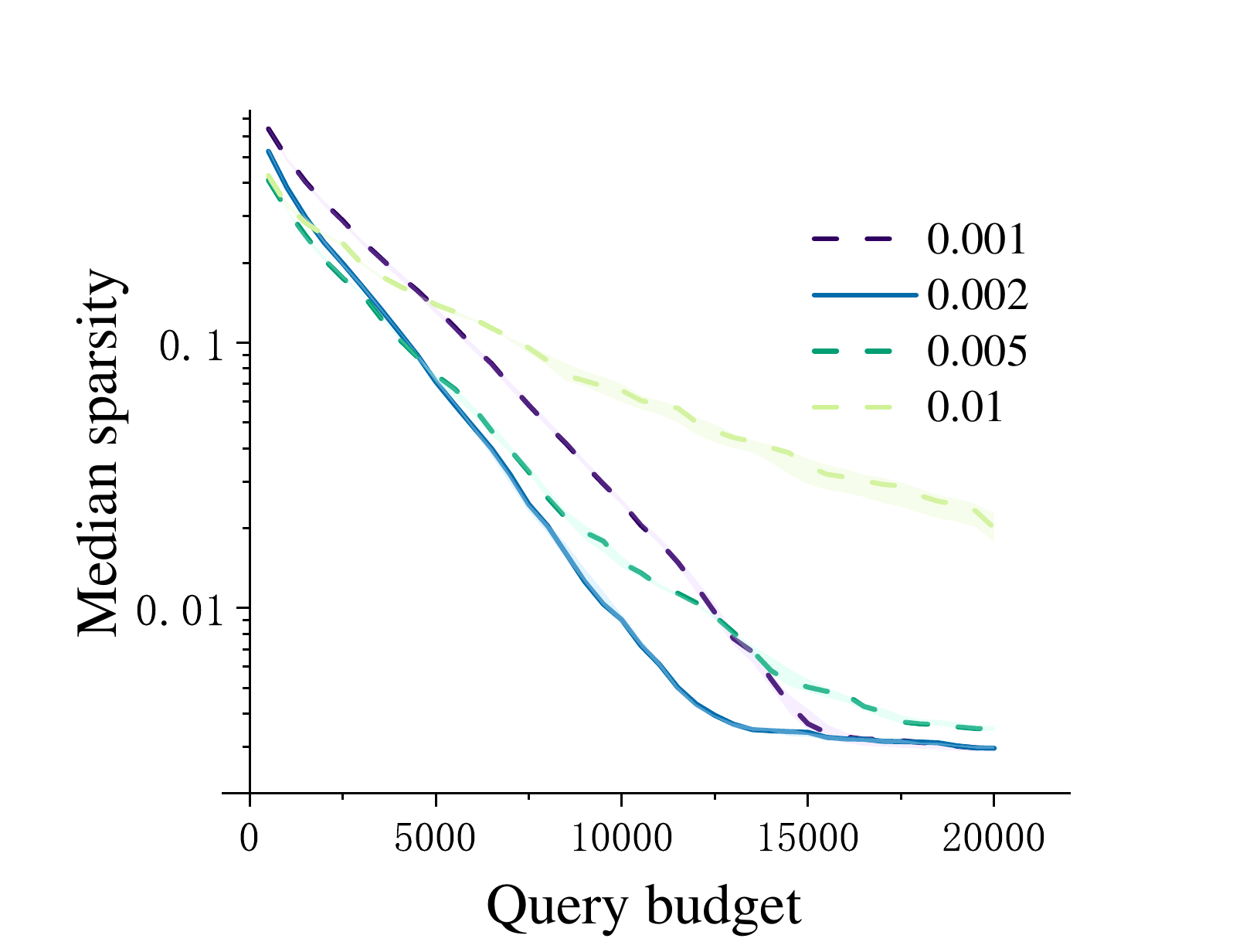}}\hfill
\subfloat[DQN learning rate.]{\includegraphics[width=0.49\linewidth]{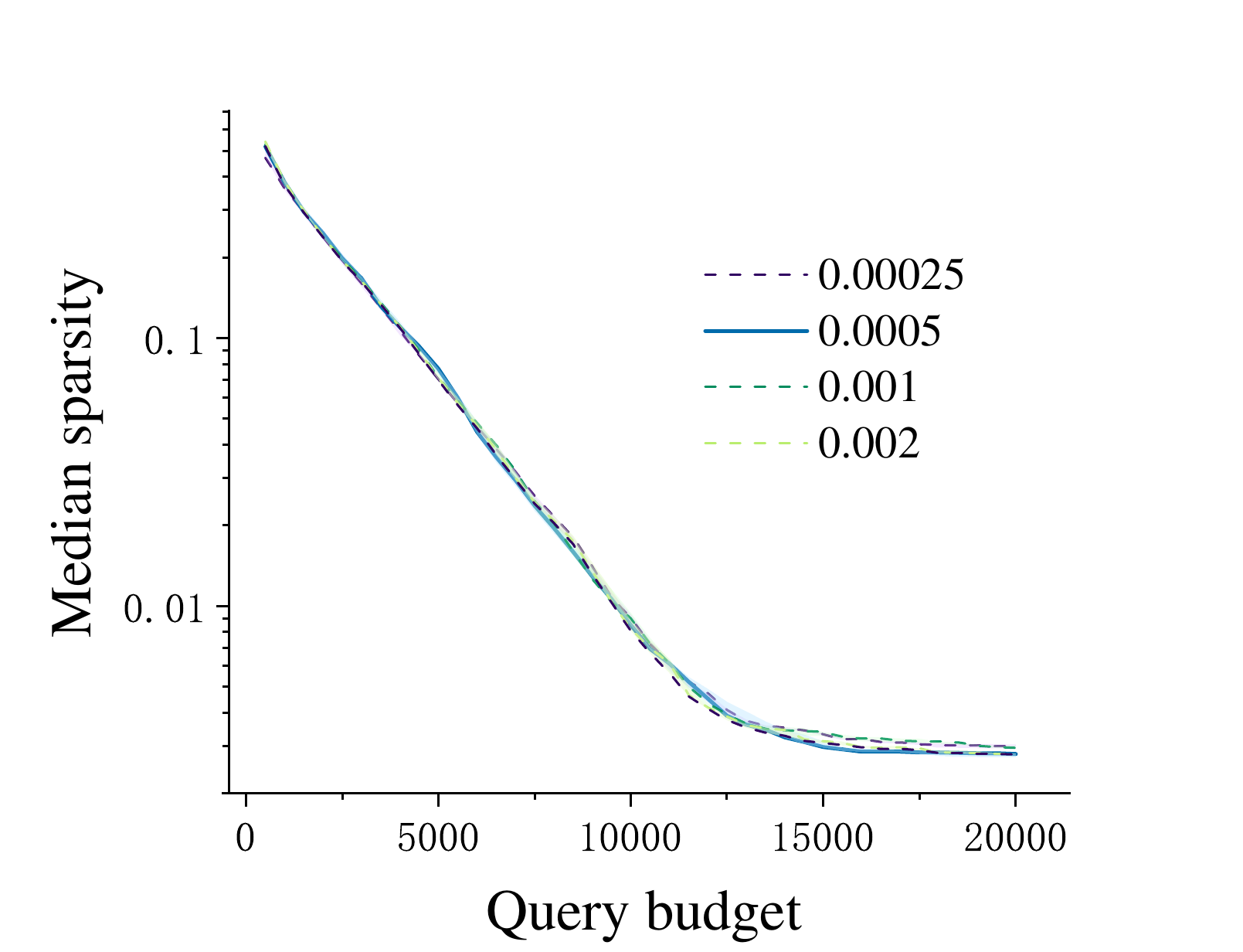}}
\caption{ImageNet targeted hyperparameter sensitivity on ResNet50. Curves aggregate three seed-specific median-sparsity trajectories over the same 100-image stream; shaded regions denote 95\% confidence intervals. The horizontal line marks the sparsity threshold 0.005 used in the main targeted comparison. Lower is better.}
\label{fig:imagenet-hparam}
\end{figure}

We assess sensitivity to hyperparameter choice by varying one parameter at a time on targeted ResNet50, using the same 100-image stream, a 20,000-query budget, and DQN seeds 42, 123, and 3407. The official configuration ($\rho=0.002$, learning rate $10^{-3}$) remains fixed; this is a post-hoc analysis, not a search that changes the reported setting. Figure~\ref{fig:imagenet-hparam} reports the geometric mean of the three median-sparsity trajectories with 95\% confidence intervals.

The relative erase ratio controls a granularity trade-off. Default $\rho=0.002$ reaches sparsity 0.005 at about 11.5K queries, earlier than $\rho=0.001$ (14.5K) and $\rho=0.005$ (15.5K). At 20K queries, these three converge to 0.00299, 0.00293, and 0.00352, respectively, while $\rho=0.01$ remains at 0.01993 and never reaches 0.005. The default is therefore chosen for fastest threshold crossing, not lowest terminal value, while avoiding the instability of overly coarse groups.

Learning-rate sensitivity is weaker. For rates from $2.5\times10^{-4}$ to $2\times10^{-3}$, all curves reach sparsity 0.005 between 11.5K and 12K queries and finish between 0.00276 and 0.00299, with overlapping confidence bands. This eightfold range indicates that ImageNet performance is not tied to a finely tuned DQN step size, supporting $10^{-3}$ as the shared default.

\section{Conclusion}

We introduced RLGPA, an online reinforcement learning-guided granularity planning attack for decision-based sparse adversarial examples. RLGPA exposes stable executor--scale actions, converts them into support-relative groups, and dynamically rebuilds those groups around a feasibility-preserving adversarial incumbent. A shared Dueling Double DQN learns from hard-label support gains during the attack stream and can be frozen for update-free evaluation. Under the matched top-1-label oracle, results on CIFAR-10 and ImageNet show strong improvements over the directly comparable Pointwise and SparseEvo attacks in the low-sparsity region. Matched controls establish that the policy learns reusable state-conditioned scheduling, while ablations identify dynamic repartition and Memory-guided late refinement as the principal mechanisms. The evidence also shows where the current method does not yet win: learned scale selection and shaped reward are not significantly superior to simpler alternatives. This combination of attack performance, causal attribution, and explicit limitations provides a reproducible basis for studying online reinforcement learning in query-bounded support search.


\bibliographystyle{IEEEtran}
\bibliography{refer}

\newpage

 




\vfill

\end{document}